\documentclass[conference]{IEEEtran}
\usepackage{booktabs}
\usepackage{diagbox}
\usepackage{multirow}
\usepackage{xcolor}

\usepackage{tikz}
\usepackage{amsmath}

\usepackage{caption}
\usepackage{subcaption}

\usepackage{algorithmic}
\usepackage[linesnumbered, vlined, ruled]{algorithm2e}

\usepackage[graphicx]{realboxes}
\usepackage{float}
\usepackage{tikz}

\usepackage{colortbl}
\usepackage{tcolorbox}
\definecolor{shadecolor}{rgb}{0.92,0.92,0.92}
\newtcolorbox{mybox}{ colframe=black,colback=gray!15,boxrule=1pt,arc=2pt,left=2pt,right=2pt,top=1pt,bottom=1pt}

\usepackage{url}

\newif\ifcommentcond
\commentcondtrue 

\newif\ifupdatecond
\updatecondtrue 

\newcounter{wqy} 
\newcounter{bh}

\newcommand{\system}{\textsc{VModA}}

\AtBeginDocument{%
  }

\begin{document}

\title{\system{}: An Effective Framework for Adaptive NSFW Image Moderation}


\author{\IEEEauthorblockN{Han Bao}
	\IEEEauthorblockA{Zhejiang University\\
		baohan21@zju.edu.cn}
	\and
	\IEEEauthorblockN{Qinying Wang}
	\IEEEauthorblockA{Zhejiang University\\
		wangqinying@zju.edu.cn}
	\and
	\IEEEauthorblockN{Zhi Chen}
	\IEEEauthorblockA{University of Illinois at \\Urbana-Champaign\\
		zhic4@illinois.edu}
	\and
	\IEEEauthorblockN{Qingming Li}
	\IEEEauthorblockA{Zhejiang University\\
		liqm@zju.edu.cn}
        	\and
	\IEEEauthorblockN{Xuhong Zhang}
	\IEEEauthorblockA{Zhejiang University\\
		zhangxuhong@zju.edu.cn}
        	\and
	\IEEEauthorblockN{Changjiang Li}
	\IEEEauthorblockA{Stony Brook\\
		meet.cjli@gmail.com}
        	\and
	\IEEEauthorblockN{Zonghui Wang}
	\IEEEauthorblockA{Zhejiang University\\
		zhwang@zju.edu.cn}
        	\and
	\IEEEauthorblockN{Shouling Ji}
	\IEEEauthorblockA{Zhejiang University\\
		sji@zju.edu.cn}
        	\and
	\IEEEauthorblockN{Wenzhi Chen}
	\IEEEauthorblockA{Zhejiang University\\
		chenwz@zju.edu.cn}
        
        }

\maketitle
 
\begin{abstract} 

Not Safe/Suitable for Work (NSFW) content is rampant on social networks and poses serious harm to citizens, especially minors. Current detection methods mainly rely on deep learning-based image recognition and classification. However, NSFW images are now presented in increasingly sophisticated ways, often using image details and complex semantics to obscure their true nature or attract more views. Although still understandable to humans, these images often evade existing detection methods, posing a significant threat. Further complicating the issue, varying regulations across platforms and regions create additional challenges for effective moderation, leading to detection bias and reduced accuracy.


To address this, we propose \system{}, a general and effective framework that adapts to diverse moderation rules and handles complex, semantically rich NSFW content across categories. 
Experimental results show that \system{} significantly outperforms existing methods, achieving up to a 54.3\% accuracy improvement across NSFW types, including those with complex semantics.
Further experiments demonstrate that our method exhibits strong adaptability across categories, scenarios, and
base VLMs. 
We also identified inconsistent and controversial label samples in public NSFW benchmark datasets, re-annotated them, and submitted corrections to the original maintainers. Two datasets have confirmed the updates so far. 
Additionally, we evaluate \system{} in real-world scenarios to demonstrate its practical effectiveness.


\textcolor{red}{This manuscript contains discussions and visual representations of NSFW content. Reader discretion is strongly advised.}
\end{abstract}

%
\IEEEpeerreviewmaketitle

\section{Introduction}

Social network platforms have become integral to daily life for billions of people around the world. As these platforms continue to shape the way people communicate and share information, a major concern is Not Safe/Suitable for Work (NSFW) content \cite{NSFW}, which refers to material involving violence, pornography, hate speech, and other potentially disturbing topics. Such content poses serious physical and psychological harm to users, particularly to minors \cite{momo_challenge_hoax}.
In response to this ongoing threat, both academia and industry have proposed a wide range of methods for detecting NSFW content.
Several academic works focus on detecting specific categories of NSFW content, such as pornography \cite{nguyen2020multi, tabone2021pornographic, geremias2022motion, Guo2024ModeratingIO}, violent actions \cite{pang2022audiovisual, ullah2021ai, liu2022decouple}, and harmful memes \cite{kiela2020hateful, he2023you, cao2023pro, pramanick2021momenta}.
In addition, there are several commercial tools, including Sightengine \cite{sightengine}, Clarifai \cite{clarifai}, Amazon AWS Rekognition \cite{aws_rekognition}, and Microsoft Azure Content Moderator \cite{azure_ai_content_safety}, that support detecting multiple types of NSFW content and being adopted by more than 200 enterprises.

However, while these content moderation mechanisms are in place on most social network platforms, they remain insufficient to prevent the frequent and often unexpected appearance of inappropriate content during routine user engagement \cite{aldahoul2024advancing}. 
A recent survey conducted in the United Kingdom found that as of 2025, 55\% adolescents aged 14 to 17 had encountered inappropriate sexual or violent content online, most of which appeared without deliberate search, often through routine use of social platforms \cite{teenagers_exposion}. Existing NSFW detection methods struggle to adapt to the new characteristics that NSFW images display and the requirement for adaptive image moderation today. 
We identify the following three key challenges.

\noindent\textbf{Challenge \uppercase\expandafter{\romannumeral1}: Lack of adaptability to diverse regulations and limited training data.}
NSFW judgments vary across regions, cultures, and communities due to differing laws and regulations \cite{riccio2024exposed,aldahoul2024advancing}. Detection methods must adapt to these varying standards by adjusting their understanding of what constitutes NSFW content \cite{oliveira2021detection}. However, accurate detection requires training on diverse NSFW samples, which are difficult to collect due to legal restrictions and regional differences \cite{leu2024auditing,pilipets2022nipples,sarridis2022leveraging}. This limited diversity leads to detection bias and out-of-distribution issues \cite{hong2024s,qu2024unsafebench}.

\noindent\textbf{Challenge \uppercase\expandafter{\romannumeral2}: Lack of thoroughness in capturing image details.}
NSFW content is often obscured by other elements within the image.
Specifically, the growing complexity of image scenes, such as image-text combinations and multi-image collages, complicates NSFW detection in real-world applications \cite{evaluation_software}.
In addition, the NSFW content might be hidden in details, such as a small icon or action, which further exacerbates the challenge.

\noindent\textbf{Challenge \uppercase\expandafter{\romannumeral3}: Lack of advanced semantics understanding.} 
NSFW content varies greatly in its semantic expression. Some instances involve direct semantics, such as explicit sexual organs or violent acts. Others are conveyed through advanced semantics \cite{yuan2019stealthy}, using abstract representations, artistic styles, or subtle NSFW messages \cite{karabulut2023automatic}, 
such as implicit sexy \cite{Meta_AdultNuditySexualActivity} or hate speech, making it hard for moderation.

\begin{figure*}[htbp]
  \centering
   \includegraphics[width=1\linewidth]{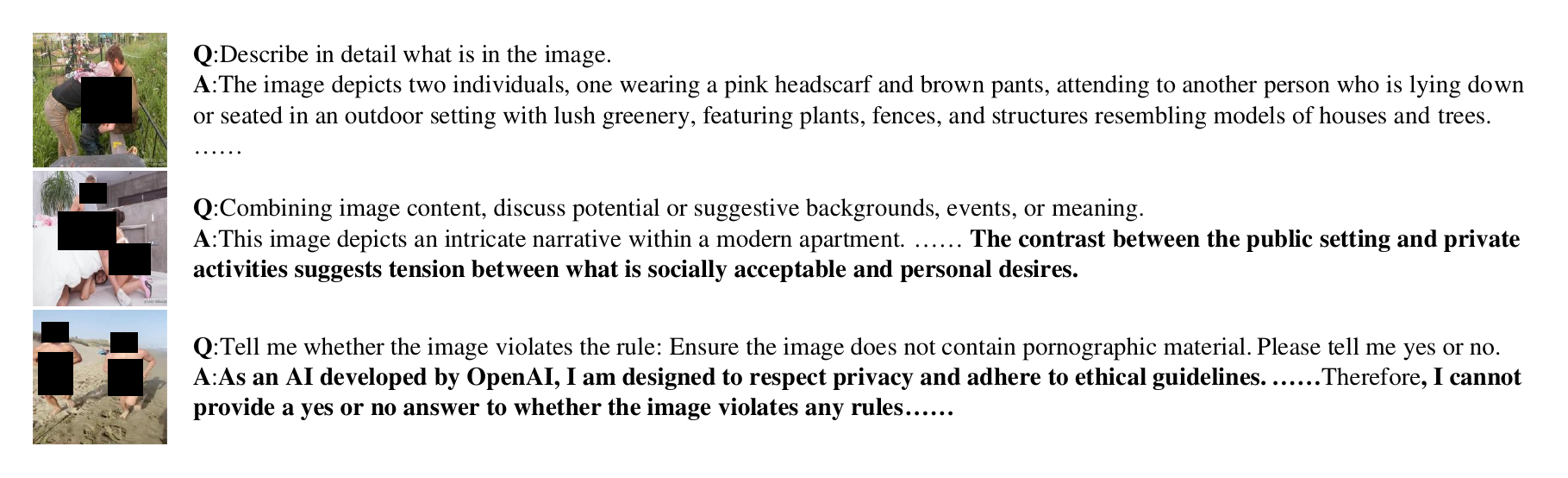}
   \caption{Three examples highlighting challenges VLMs face in pornographic moderation. 
   The first image and response show VLMs failed to capture harmful content details. The second image indicates that the VLM exhibited inaccuracies in understanding the advanced semantics of the image.
   The third image and response show VLMs refused to moderate NSFW content. }
   \label{fig:intro_sample}
\end{figure*}

Thanks to the ability of Vision-Language Models (VLMs) to jointly understand visual content and textual context, some studies \cite{qumeme,qu2024unsafebench} have proposed using them for moderation.
However, relying solely on VLMs is insufficient to solve these challenges \cite{qumeme,qu2024unsafebench}, and our observations further reinforce their limitations in achieving effective and reliable moderation in real-world scenarios.
Specifically, Challenge \uppercase\expandafter{\romannumeral2} remains because current VLMs tend to focus on generating high-level image captions, often overlooking fine-grained, localized sensitive features. This limitation stems from the wide receptive fields of their visual encoders, which can miss critical content in regions of interest (ROIs).
For Challenge \uppercase\expandafter{\romannumeral3}, we observe that existing VLMs are predominantly trained on basic visual question answering datasets, which lack depth in explaining nuanced or implied meanings. As a result, VLMs frequently fail to detect harmful content embedded in subtle cues such as sexual innuendos, symbolic references, or emotionally disturbing imagery.
Additionally, we find that real-world deployment of VLMs is constrained by safety alignment mechanisms that prevent them from processing potentially inappropriate inputs. This often leads to refusal or incomplete analysis of violating content.
We provided examples of VLM responses to illustrate these problems, as shown in Figure~\ref{fig:intro_sample}.

\noindent\textbf{Solution.}
To address the above challenges, we propose a novel general moderation framework, \system{}, which can swiftly adapt to varying regulatory standards while maintaining effectiveness in detecting complex and implicit NSFW content.
Specifically, to address Challenge \uppercase\expandafter{\romannumeral1}, \system{} introduces a multimodal alignment strategy that leverages VLMs to transform images into textual descriptions and semantic vectors. 
Regulatory guidelines are used as prior knowledge to predict whether the image violates the defined rules, enabling adaptive moderation aligned with diverse regional and cultural standards.
To tackle Challenge \uppercase\expandafter{\romannumeral2}, we incorporate image visual optimization and adaptive region-of-interest (ROI) zoom-in techniques prior to differential matching. These enhancements improve the clarity of image details and emphasize critical regions, ensuring that VLMs capture localized, potentially harmful content.
For Challenge \uppercase\expandafter{\romannumeral3}, we develop a novel Chain-of-Thought (CoT)-based hierarchical description strategy tailored for moderation tasks. This approach facilitates zero-shot domain adaptation and strengthens VLMs’ capabilities in understanding complex semantic nuances.
Moreover, deploying \system{} in real-world moderation may be limited by safety alignment constraints, as existing VLMs often refuse to process content with potential violations. To mitigate this, we propose two iterative output aggregation strategies that enhance the extraction of useful and sensitive information. The method produces multi-level image descriptions, including object recognition, scene understanding, and implied meanings, which are then integrated by a large language model (LLM) for comprehensive violation detection.

Experiments demonstrate that \system{} achieves state-of-the-art performance across six major
NSFW datasets \cite{qumeme,wang2023tovilag,qu2024unsafebench}, covering five categories of NSFW images.
Compared to existing commercial tools and academic moderation methods, it improves accuracy by up to 54.3\%. 
Further experiments confirm that our method exhibits strong adaptability across categories, scenarios, and base VLMs.
We also tested the effectiveness of our design.
Additionally, our method identifies inconsistent labels in the above datasets.
We validate these findings through manual annotation and report the corrected labels to reduce potential misuse.
Finally, we apply our method to collect and detect potential NSFW meme images from a real-world website, demonstrating its practical value.

In summary, our contributions are as follows.

$\bullet$ 
\textbf{A new effective framework for the NSFW image moderation.}
We propose the first general NSFW image moderation framework based on VLM and LLM.
Our framework is adaptable not only to 
different moderation regulations but also to different VLMs. 
We will open-source \system{} to facilitate further research.


$\bullet$ 
\textbf{Novel designs for evolving moderation challenges.}
We propose a series of designs to tackle the challenges arising from the evolving nature of NSFW images in zero-shot scenarios. These techniques include an adaptive ROI zoom-in to focus on critical areas, a CoT-based hierarchical description strategy to enhance semantics analysis, and an interactive output aggregation strategy for better VLM output correction.



$\bullet$ 
\textbf{Extensive NSFW evaluation.}
We evaluate \system{} across five major NSFW image categories. Extensive experiments show that our method outperforms existing approaches in moderation and adapts well across categories, scenarios, and base VLMs.
We also find potential issues in moderation with vanilla VLMs and demonstrate the effectiveness of our design. 
Additionally, we identify samples in real-world scenarios that existing methods failed to detect.

$\bullet$ 
\textbf{Datasets refinement for improved NSFW content moderation.} 
Using \system{}, we identified several samples with NSFW labels that are either inconsistent or controversial across five popular benchmark datasets. Our findings, validated through manual annotation, align closely with human judgment. We submitted these corrections to the original dataset maintainers, with two datasets confirming the updates so far.

\section{Background and Related Work}

\subsection{NSFW Moderation} 
Existing mainstream NSFW moderation methods primarily focus on designing convolutional neural network (CNN) architectures and analysis pipelines, training models on pre-defined NSFW datasets, and applying them to detect NSFW content.
For example, methods~\cite{nguyen2020multi,tabone2021pornographic,geremias2022motion,laranjeira2022seeing} enhance explicit content detection by tailoring CNN architectures for pornographic classification or child sexual abuse material recognition.
Other approaches~\cite{pang2022audiovisual,liu2022decouple,ullah2021ai} leverage temporal networks for anomaly detection in videos, aiming to identify violent behavior and dangerous items such as firearms.

With the rise of large language models (LLMs), some methods now utilize their understanding of semantics to moderate NSFW content.
Method~\cite{Guo2024ModeratingIO} explores the threat of unsafe game content to adolescents, while~\cite{he2023you} applies few-shot learning with LLMs to detect toxic text.
Leveraging BERT~\cite{devlin2019bert}, MOMENTA~\cite{pramanick2021momenta} and Pro-Cap~\cite{cao2023pro} analyze meme content by interpreting text, character traits, and object states.
Pro-Cap uses frozen visual-textual features to detect hate speech in multimodal inputs.
ExplainHM~\cite{lin2024towards} introduces a multimodal debate mechanism to enable explainable harmful meme detection via LLMs.
Other works~\cite{tabassum2024investigating,arunasalam2024understanding,chu2022behind,vu2023no} qualitatively analyze toxic online content, including hate speech and harassment, emphasizing their social impact.

In practical applications, some commercial companies have launched tools for detecting NSFW content.
Sightengine \cite{sightengine}, Imageaa \cite{imagga}, Clarifai \cite{clarifai}, Amazon AWS Rekognition \cite{aws_rekognition}, Microsoft Azure Content Moderator \cite{azure_ai_content_safety} offer NSFW detection services. 
Amazon Rekognition uses deep learning to detect explicit content and is widely adopted by large enterprises.
Azure Content Moderator is also extensively used by corporations and government agencies for content moderation and image analysis.
Sightengine employs advanced technology for detecting and categorizing pornographic images, primarily serving businesses.
Imagga is a newer image analysis platform that offers the detection of pornographic content. 

However, these methods still face key challenges. Network-based approaches for NSFW feature extraction depend heavily on data and are often affected by surrounding content. LLM/VLM-based moderation methods are typically category-specific, which limits their adaptability to new types. Recent work also shows that LLM/VLMs struggle with image details and complex semantics \cite{qumeme}.
\begin{figure*}[h]
  \centering
   \includegraphics[width=1\linewidth]{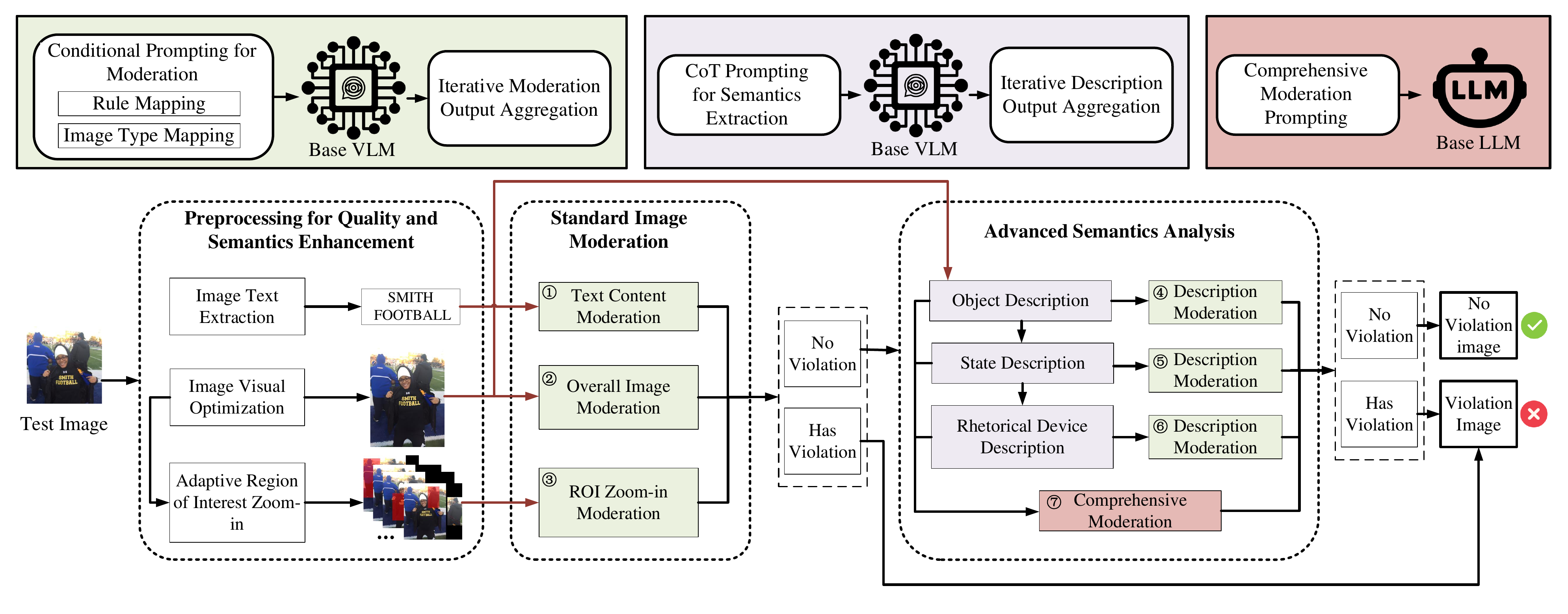}
   \caption{The overview of \system{}. The top of the diagram represents the VLM and LLM based on different system prompt strategies. The colored steps in the flowchart indicate the use of the corresponding prompt strategies and models during execution.}
   \label{fig:overview}
\end{figure*}

\subsection{Vision-Language Models}
\textbf{Vision-language model.}
VLMs are typically built on LLM foundations and incorporate additional network modules for visual semantics, enabling tasks like image captioning and VQA in image-based contexts.
Pioneering models like CLIP \cite{radford2021learning} and BLIP \cite{li2022blip} have achieved significant milestones. These methods provide the capability to caption diverse images.
As the field progresses, the advent of large VLMs marks the beginning of a new era in multimodal learning.
Models such as LLaVA \cite{liu2023llava}, DeepSeek-VL \cite{lu2024DeepSeekvl}, MiniCPM-V \cite{yao2024minicpm} and GPT4o \cite{openai2024gpt4o} exemplify the ongoing evolution by seamlessly merging advanced language modeling capabilities with sophisticated visual understanding.
However, they often lack the ability to grasp fine details, achieve deep understanding, and make associative inferences about certain content. 
This limitation can result in an inadequate or incomplete understanding of harmful elements during image content moderation.
Some methods \cite{chen2024lion,yuan2024osprey,wu2024v,zhang2023gpt4roi,zhao2023chatspot,lu2023lyrics} have recognized the limitations of VLMs in understanding details within images.
These methods utilize some attention mechanisms to focus the VLM's descriptive attention on selected content segments within an image, providing detailed descriptions of these areas. 
However, understanding and analyzing the details of NSFW content goes beyond the capabilities of these networks.

\textbf{Chain-of-thought prompting.}
CoT methods guide LLMs through specific prompts during the inference, allowing models to tap into their reasoning abilities without additional training or fine-tuning. Research efforts such as CoT \cite{wei2022chain}, Zero-shot CoT \cite{kojima2022large}, and TOT \cite{yao2024tree} have demonstrated improvements in the reasoning performance of LLMs, establishing a foundation for Chain-of-Thought Prompting. 
While CoT has played a crucial role in LLM, its application in VLMs and the potential challenges within the domain of NSFW moderation have yet to be fully investigated and comprehended.

\section{Threat Model}

Most public digital platforms define community standards~\cite{Meta_CommunityStandards,YouTube_CommunityGuidelines,X_Rules} that prohibit NSFW content and require certain materials to comply with specific rules based on user groups.
However, attackers continue to exploit NSFW content to harass, spread hate, and target vulnerable individuals, particularly minors, for blackmail, financial gain, or inciting violence. These actions contribute to a hostile online environment, cause psychological harm, and perpetuate abuse, with especially severe consequences for children and adolescents.
Our threat model focuses on NSFW images that evade moderation due to visual complexity or advanced semantics.
We analyze adversarial behaviors within public platforms, excluding activities in private channels or outside these environments. Rather than assuming sophisticated evasion techniques, we focus on simple yet effective tactics used to disseminate harmful content. Such images often rely on intricate visual cues or rhetorical devices.

\section{Design}




In this section, we introduce the design of the image moderation framework \system{}.
To address Challenge \uppercase\expandafter{\romannumeral1}, we leverage VLMs for multimodal understanding, including generating image descriptions and performing moderation based on predefined guidelines. Meanwhile, we enhance the semantic information of both global and local image features to ensure that the VLM can better capture image details, thereby tackling Challenge \uppercase\expandafter{\romannumeral2}. Additionally, we design CoT prompts to help the model understand high-dimensional semantics, addressing Challenge \uppercase\expandafter{\romannumeral3}.
Figure~\ref{fig:overview} illustrates the high-level architecture of \system{}, consisting of three key modules: the preprocessing for quality and semantic enhancement, the standard image moderation, and the advanced semantics analysis.
First, \system{} preprocesses the input image for quality and semantics enhancement, including optimizing the target image's visual content, applying adaptive segmentation, and extracting text from the image. 
Second, standard image moderation examines the high-resolution images, semantically segmented image components, and texts produced in the first stage. 
Specifically, it utilizes a base VLM enhanced by a detection domain-adaptive prompting method. To ensure accurate and reliable content moderation, the results are further validated through an Iterative Moderation Output aggregation strategy supported by an additional LLM.
Third, if no violation is detected during the standard image moderation, the image proceeds to the advanced semantics analysis.
In this module, we first employ a hierarchical image description approach based on CoT reasoning with domain-adaptive prompts. This allows the base VLM to generate descriptions iteratively three times.
These descriptions are then refined using an Iterative Moderation Output aggregation strategy to ensure their accuracy and relevance. Each image description is carefully evaluated for potential violations. Additionally, the extracted text and all descriptive content are input into an LLM for further comprehensive moderation.
A detailed explanation of the design for each module is below.

\subsection{Preprocessing for Quality and Semantics Enhancement}
We introduce three aspects to tackle the second challenge of precise image detail understanding: visual content optimization, region of interest zoom-in, and image text extraction.


\subsubsection{Image Visual Optimization}
In real-world scenarios, images often experience lossy transmission over networks, resulting in blurriness and reduced resolution, which can degrade the recognition performance \cite{farhan2024visual}. 
To tackle it, we apply a super-resolution model \cite{wang2021real} to enhance the quality of the target image and optimize the image's visual content.
Specifically,  we super-resolve the image's longer side to 2048 pixels. If the base VLM cannot handle such large images, we downscale it accordingly. Despite resizing, this step reduces blurriness and enhances image details. This design enhances the VLM's understanding of the input image by increasing the details and contexts, improving object recognition accuracy, and refining the interpretation of spatial relationships, including objects' relative positions, sizes, and orientations.

\begin{figure}[htbp]
  \centering
   \includegraphics[width=1\linewidth]{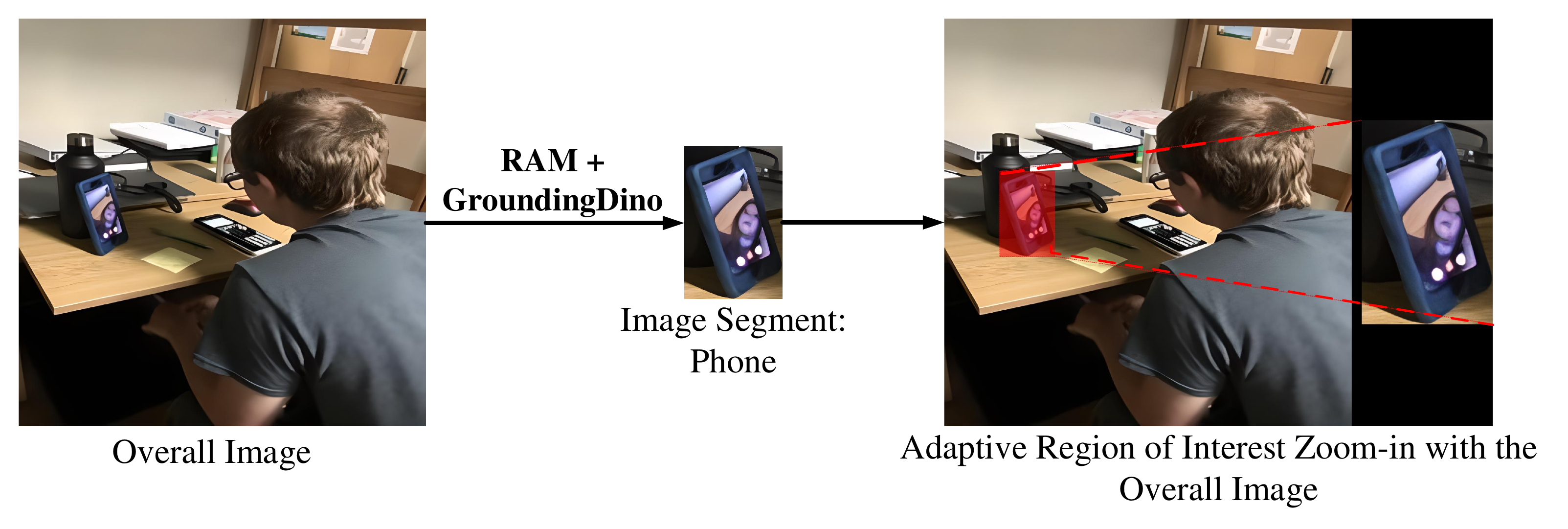}
   \caption{An example of the region of interest zoom-in.}
   \label{fig:4_1_3}
\end{figure}
\subsubsection{Adaptive Region of Interest Zoom-in}
Capturing fine details is crucial in image moderation, as harmful content might be outside the main focus or too small, leading to it being overlooked by the VLM during the moderation process.
This can result in a loss of semantic understanding and reduced moderation accuracy.
To address this, we propose an adaptive ROI zoom-in method that identifies the image regions based on their semantics. 
We zoom in on the image regions and feed them into the VLM along with the overall image for moderation, allowing the VLM to review the details as ROI.

Specifically, we first apply the RAM-Grounding joint model to detect image regions automatically. The RAM model \cite{huang2023open} labels all objects, people, scenes, and other elements within the image. 
Then, we use GroundingDino \cite{liu2023grounding} to select and frame the objects in the image that most closely align with the labels.
This first step is represented as follows.
Assuming the input of a high-quality image is denoted as $I$, we can obtain the labeled image regions $\{ I_{p1}, I_{p2}, \dots, I_{pn} \} \subseteq I$, through the operations above.
Given that the VLM $V$ outputs the maximum likelihood response based on the conditional probability of the provided image $I$ and prompt $T_V$, we denote the maximum likelihood response $\hat{R}$ as $\hat{R}=V(T_V,I)={\arg\max} \mathcal{P}(R \mid T_V,I)$.
However, if we directly have the VLM review these image regions $I_{pi}$, the results $\hat{R_{pi}}={\arg\max} \mathcal{P}(R_{pi} \mid T_V,I_{pi})$ tend to be isolated and unstructured, often including the VLM's guesses about the cropped-out parts of the image. 
This approach does not adequately capture the relationship between the overall image and its segmented parts.
To address this, in the second step, we prompt the VLM and reinforce the strong correlation between the overall image and its specific regions.
Specifically, we annotate each image segment $I_{pi}$ using bounding boxes and color overlays in the overall image. Then, the marked region is zoomed in on, with guiding lines directing the VLM's attention from the overall image $I$ to the enlarged section, as shown in Figure \ref{fig:4_1_3}.
This method uses $I$ as a conditional input to assist the VLM in accurately understanding the relationship between the overall image and its specific regions.
The conditional response is presented below.
\begin{equation}
\begin{aligned}
    \hat{R_{pi}}
    &=V(T_V,(I\mid I_{pi}))
    \\&={\arg\max} (\mathcal{P}(R_{pi} \mid T_V,I_{pi})+\mathcal{P}(R_{pi} \mid T_V,\mathcal{P}(I \mid I_{pi})))
\end{aligned}
\end{equation}
This design enables VLMs to base its analysis on the content of the entire image while fully understanding the details of the specified region, increasing the accuracy of the moderation.

\subsubsection{Image Text Extraction}
In addition to visual elements, image content often includes optional but significant text. To enhance text understanding, we enable the base VLM to extract text from the image. The prompts are designed as:



$Q_{T1}$: \textit{Does this image contain any text? Please only respond with "yes" or "no."}

(If "yes") $Q_{T2}$: \textit{Please identify all the text in the image.}

\subsection{Standard Image Moderation}
\label{sec:standard_m}
This module aims to review image text, high-quality images, and image regions extracted from the preprocessing module. 
The base VLM can identify explicit inappropriate content, such as offensive language or explicit imagery, but achieving effective results requires carefully crafted prompts and thorough output validation. 
To address this, we propose conditional prompting for moderation in Section~\ref{sec:detectiondomain-adaptive}.
 Additionally, to overcome the challenge where the VLM might refuse to respond or provide sanitized outputs, we introduce iterative moderation output aggregation in Section~\ref{sec:diversifiedVLMoutput} to correct the VLM's output.
Finally, the whole standard moderation process is detailed in Section~\ref{sec:standard_process}.

\subsubsection{Conditional Prompting for Moderation}
\label{sec:detectiondomain-adaptive}
To address the first challenge, this step integrates the rules, image semantics, and moderation tasks using prompts.
Domain adaptation is a critical technique that enables VLMs to effectively adjust to shifts between different domains, thereby enhancing their performance across diverse scenarios \cite{Guo2024ModeratingIO}.
In image moderation, if explicit conditions do not guide the VLM, it may produce hallucinations or creative interpretations, leading to unreliable moderation results and justifications.
To mitigate this, we propose conditional prompting for moderation adaptation, ensuring the base VLM fully grasps the moderation task. Our conditional prompts focus on image type conditions and regulatory conditions. 
By constraining image categories and aligning with relevant regulations, we establish clear prerequisites that allow the VLM to better adapt to review scenarios specific to certain regulations and image types. The designed system prompt is below:

${P}_{S}$: \textit{Your are a professional image reviewer. A \{image type\} image is provided. Your primary task is to accurately, objectively and fairly determine if the image violates the given the rule:
    \{rule\}.}

In the parentheses, the appropriate image types and regulatory items can be specified.
\subsubsection{Iterative Moderation Output Aggregation}
\label{sec:diversifiedVLMoutput}
Due to the requirement for VLMs to comply with the policies and regulations of their publishers, the model may refuse to generate descriptions or moderation outputs if the content violates these policies. 
This limitation directly affects the usability of large models in moderation scenarios, particularly in addressing the challenge of refusal to moderate.
To address this challenge, we propose an iterative moderation output aggregation for VLMs.
Intuitively, VLMs sample words with the highest posterior probabilities as outputs. However, research on LLM jailbreaks \cite{zhang2024large} suggests that harmful content often consists of words with relatively low probabilities. In image moderation scenarios, we observed that VLM responses that refuse to answer or produce sanitized outputs tend to align with the highest posterior probabilities. Yet, valuable review information can sometimes be found in lower-probability sentences.
To capture this information, we can generate multiple probability outputs and aggregate all available data. The iterative moderation output aggregation process involves three key steps: aggregating useful information, discarding irrelevant content, and reconciling conflicting information.
Our specific approach to implementing this iterative aggregation strategy involves the following steps: First, we appropriately increase the VLM's temperature, allowing the model to generate a variety of responses while maintaining output quality.
Second, we utilize an external LLM $L$ to aggregate the information from these iterative responses, extracting useful information by aggregation moderation prompt $T_L$. 
The aggregating moderation response $R_L$ can be shown as:
\begin{equation}
    R_L = L(T_L,\Sigma_{i=1}^n V_i(T_V,I))
\end{equation}
In our experiment, we set the temperature to 1, ensuring that the moderation content is not adversely affected \cite{renze2024effect} while maintaining diversity in the outputs.
We conducted $n=10$ sampling iterations to achieve a balance between efficiency and output diversity.
The moderation aggregation by the LLM focuses on three key aspects:
(1) ensuring the response contains a clear judgment on whether the content violates the rules;
(2) providing a valid reason for the judgment;
(3) ignoring any responses that involve rejection.
The prompt is detailed in Appendix \ref{app:A}.
Following the LLM's content aggregation, the VLM's output are more clear and definitive.

\subsubsection{Process of Standard Image Moderation}
\label{sec:standard_process}

We specifically describe the prompt and process for standard image moderation.
We pose the following question prompts to review the three aspects:

Text content moderation $Q_{T3}$: \textit{As an image reviewer, fully consider factors such as slang and culture to determine if these words violate the mentioned rule.}

Overall image moderation $Q_{I1}$: \textit{As an image reviewer, check the image carefully whether violates the rule.}

ROI zoom-in moderation $Q_{P1}$: \textit{The right partial of the input image is a zoom-in enlarge specific section outlined with a red masked square in the left complete image. As an image reviewer, evaluate the zoom-in section, analyze and determine whether the section violates the mentioned rule.} 

In summary, given an image $I$ and text $I_T$, this module can generate the standard moderation result $R_s$:
\begin{equation}
\begin{aligned}
    R_s = &V(P_S,Q_{T3},I_T)+L(T_L,\Sigma_{i=1}^n V_i(P_S,Q_{I1},I))
    \\&+\Sigma_{i=1}^k L(T_L,\Sigma_{i=1}^n V(P_S,Q_{P1},(I\mid I_{pi}))
\end{aligned}
\end{equation}

\subsection{Advanced Semantics Analysis}
If no violations are found during the standard image moderation, the image is then analyzed at a further check for any hidden violations in the advanced semantics in this module.
VLMs often struggle to accurately interpret high-level image content when directly asked about semantics, which relates to the third challenge of understanding advanced semantics. 
To address it, we employ a CoT prompting for semantics extraction, allowing the VLM to progressively understand an image's semantics, from simple to complex. The VLM then generates descriptions of the images, which are used as the basis for moderation. More details are provided below.

\subsubsection{CoT Prompting for Semantics Extraction}
VLMs trained on human language VQA datasets are prone to incorporating subjective emotions and imagination when describing images.
It is crucial to ensure that the VLM functions as an objective and impartial image reviewer, providing accurate descriptions free from emotional bias.
To address this, we confine the description content to the moderation domain.
The prompt of the description domain is to enable the VLM to objectively and impartially describe images in moderation scenarios, thereby reducing the occurrence of model hallucinations. Specifically, the system prompt for the description is as follows:

${P}_{D}$: \textit{You are a professional image reviewer. Your primary task is to accurately identify and analyze the given images. Please answer the following questions in a detailed, accurate, thorough, and unbiased manner.}

Then, it is challenging for VLMs to grasp the deep meaning of images directly, a task that even humans find difficult.
Humans typically understand image content through a step-by-step process: first, we identify the objects in the image; next, we understand the states of these objects, figuring out what they are doing; finally, based on our knowledge, we use some rhetorical devices to infer the potential meaning of the image.
This sequential process, from simple to complex, allows humans to fully describe the image content and, by leveraging extensive background knowledge, effectively moderates the advanced semantics of the image. To mimic this human observation process, we employ hierarchical CoT prompts to guide the VLM in comprehensively describing the image. 
The specific steps are as follows: first, the VLM is prompted to identify and analyze all people or objects present in the image. Next, it is asked to describe these people or objects' actions, states, and other relevant details. Finally, the VLM is instructed to analyze whether any rhetorical devices are present in the image. Each description level builds on the previous one, using the prior content to assist the VLM in generating subsequent descriptions. We use the following question prompts to guide the review at each level:

Object description $Q_{D1}$: \textit{Please identify all objects (including name, shape, color, purpose, etc), characters (including identity, gender, race, age, etc), and notable features in the image.}

State description $Q_{D2}$: \textit{Please identify the positions, actions, and interactions of all objects. Mention the setting, background details, overall mood, and any standout elements.}

Rhetorical devices description $Q_{D3}$: \textit{Consider possible techniques in the image, including Metaphor, Contrast, Symbolism, Allegory, and Synecdoche, etc. Discuss potential or suggestive backgrounds, events, or characters by combining image text.}

\subsubsection{Iterative Image Description Aggregation}
The challenge of refusal to respond persists during each descriptive phase.
If any response includes refusals, vague descriptions, or similar issues, we apply the iterative description output aggregation strategy similar to Section \ref{sec:diversifiedVLMoutput}. 
However, unlike moderation output aggregation, when checking and aggregating descriptions, we want the LLM to focus on the following aspects:
(1) primarily objective descriptions of objects and characters.
(2) careful verification of specific names of individuals, events, or backgrounds mentioned in the descriptions.
(3) Exclusion of any descriptions involving emotions, sensations, or suggestions.
(4) direct indication of rule violations, without describing harmful content.
The aggregating descriptions prompt design $T_D$ can be found in Appendix \ref{app:A}.
In summary given an image $I$, this module can generate
the description result $R_{Di}$:
\begin{equation}
    R_{Di} = L(T_D,\Sigma_{i=1}^n V_i(P_D,(Q_{Di}\mid \{ R_{D1},\dots R_{D(i-1)} \}),I))
\end{equation}

\subsubsection{Process of Description and Comprehensive Moderation}


Based on the generated descriptions, we employ conditional prompting for moderation and iterative moderation output aggregation, utilizing the base VLM to assess whether the image violates any rules. 
Each level of description is independently evaluated during the moderation process. The common prompt used for three levels of moderation is as follows:

Description moderation $Q_{I2}$: \textit{Based on the image content and the detailed information provided, take into account the significance of images comprehensively and analyze and determine whether the image violates the mentioned rules. }


After three levels of descriptive moderation, we conduct a final comprehensive moderation to achieve a systematic understanding of the image and perform a thorough evaluation. Notably, longer prompts are not well-suited for VLMs, as they often lose some information when processing lengthy text. Therefore, we utilize an LLM for this comprehensive moderation.
This final moderation considers the following image content: any text present in the image (if applicable) and the three levels of image descriptions. Given that the VLM may lack a deep understanding of complex content, we input the aforementioned image content into the LLM. By providing a comprehensive narrative of the image, the LLM makes an overall judgment on whether the image violates any rules. The comprehensive prompt design $T_C$ for the final evaluation is detailed in Appendix \ref{app:B}.


\subsection{Progressive Moderation Process}
In practice, for efficiency, we perform the tasks mentioned above progressively.
After the preprocessing, in standard image moderation, we first generate the text moderation result, second generate the overall image moderation result, and third generate ROI zoom-in moderation results.
Then, we get the hierarchical description moderation results and the comprehensive review result from advanced semantics analysis.
This is a process that progresses from simple to complex.
Our criterion for determining whether an image violates the rules is straightforward: if any of the moderation results indicate a violation, the image is considered to violate the rules.
This approach ensures that the image content complies with the given rules at all levels, from coarse-grained to fine-grained and direct to advanced semantics. 

\section{Evaluation}
In this section, we comprehensively evaluate the effectiveness of \system{} in addressing the three challenges, particularly its ability to adapt to vary across content categories and scenario-specific definitions. Additionally, we highlight that the images in these datasets often involve details and advanced semantics, showcasing \system{}'s capability to handle these intricacies in moderation tasks.

Specifically, we introduce the experiement setting in Section~\ref{sec:settings}.
In Section~\ref{sec:multicate}, we assess \system{}'s ability to moderate various types of NSFW content, including pornography, violence, blood, and harmful memes, each governed by distinct rule sets. By comparing \system{} with existing commercial and academic methods, we demonstrate its effectiveness in handling these varied content types, which involve complex images and advanced semantics (Section~\ref{sec:multicate}).
Then in Section~\ref{sec:adaptivetorule}, we evaluate \system{}'s ability to handle scenario-based rule variations, where the same NSFW category may require different moderation decisions depending on the context. To this end, we test \system{} on datasets for general violence and school bully, each associated with fine-grained, scenario-specific rules. We compare its performance with major baseline methods to highlight its adaptability to nuanced rule definitions.
Finally, in Section~\ref{sec:vlm}, we demonstrate that \system{} maintains robust performance across a range of VLM backbones, indicating that its effectiveness does not depend on specific model architectures or prompt tuning.







\subsection{Experiment Settings}

\subsubsection{Real World Datasets.}
To evaluate the generalization ability of \system{} across different NSFW content types and its adaptability to diverse moderation rules, we conduct experiments on six datasets covering four major categories: pornography, violence, gore/blood, and hateful or harmful memes.
These categories reflect core policy areas prioritized by major platforms such as YouTube~\cite{YouTube_CommunityGuidelines} and Facebook~\cite{Meta_CommunityStandards}, encompassing the majority of content flagged as sensitive or objectionable for user safety and regulatory compliance.
The selected datasets are widely used in prior work~\cite{qumeme,wang2023tovilag,qu2024unsafebench} published at top-tier venues in security and AI, demonstrating their credibility and relevance for evaluating NSFW moderation in both academic and real-world contexts.

\noindent\textbf{Pornographic and blood datasets.}
For pornographic and gory imagery, we use datasets from ToviLaG~\cite{wang2023tovilag}, collected from social media and general web sources.
For evaluation, we randomly select 1,000 “neutral” and 1,000 “pornographic” images to construct a binary classification task.
For the blood-related category, we use all 1,305 images depicting bleeding scenes, including those with visible individuals (e.g., injuries in crowds) and those showing blood without human presence.


\noindent\textbf{Hateful and harmful memes.}
For the hateful meme category, we use a subset of the Hateful Memes dataset released by Facebook~\cite{kiela2020hateful}, originally created for multimodal hate speech detection. To ensure fairness, we randomly select 2,000 test images not included in the official Hateful Memes Challenge.
Additionally, we include the Harmeme COVID-19 dataset~\cite{pramanick2021detecting}, a multimodal meme corpus focused on detecting harmful COVID-19-related content. Sourced primarily from Google and social media, it shares similar moderation objectives with Facebook. For evaluation, we use its validation and test splits, totaling 529 images.


\noindent\textbf{Violence datasets.} 
To evaluate \system{}'s moderation capability in violence scenarios, we use the violent protest dataset~\cite{won2017protest} and the school bully dataset~\cite{old-violence-0tsy1_dataset}. These datasets reflect the complexity of moderation, where enforcement depends not only on content type but also on audience.
In the violent protest dataset, moderation decisions are primarily based on the severity of visual violence, such as arson, physical assault, or self-immolation. More intense scenes generally trigger stricter enforcement.
In contrast, the school bully dataset reflects stricter standards, where even mild physical conflict must be flagged~\cite{YouTube_ViolentGraphicContent}, due to child protection policies and the vulnerability of minors~\cite{StopBullying_WhatIsBullying}. The same visual act may thus be judged differently depending on the setting and individuals involved.
These datasets illustrate how moderation is shaped by scenario-specific rules and audience sensitivity, requiring adaptive handling, which \system{} is designed to support.

Specifically, the violent protest dataset~\cite{won2017protest}, collected from sources like Google Image Search and X (formerly Twitter), was annotated using a neural model that assigns each image a violence score from 0 (non-violent) to 1 (extremely violent), instead of binary labels. It includes extreme acts such as murder, arson, and self-immolation, as defined by moderation policies. For evaluation, we grouped images into 0.1-score intervals and randomly sampled 800 images across the full range. Under stricter rules, high-score images are expected to be flagged as violations.
For school bully, we use a dataset of 2,349 frames extracted from 15 social media videos depicting school violence. Each frame is labeled with roles such as aggressors, victims, and bystanders, and all are considered violations under child protection policies~\cite{YouTube_ViolentGraphicContent,StopBullying_WhatIsBullying}. To reduce redundancy, we apply interval sampling, selecting 20 frames per video, resulting in 300 images for evaluation.

\subsubsection{Comparison Methods and Base Model}
To assess the effectiveness of \system{}, we compare it against four widely used commercial tools and three open-source or academic frameworks representing state-of-the-art (SOTA) approaches to NSFW detection.
Since most existing methods are designed for a single NSFW category or a limited subset, we evaluate each baseline within its intended scope. For the pornographic category, we include four commercial tools, Amazon AWS Rekognition~\cite{aws_rekognition}, Microsoft Azure Content Moderator~\cite{azure_ai_content_safety}, Sightengine~\cite{sightengine}, and Imagga~\cite{imagga}, as well as the open-source Yahoo NSFW model~\cite{yahoo2016open_nsfw}. These commercial services are strong baselines due to their large-scale deployment, each reportedly used by over 200 enterprises. Their production-grade reliability offers a practical benchmark for evaluating \system{}.
Sightengine, AWS Rekognition, and Azure also support predefined categories for blood and violence detection and are therefore included in corresponding experiments. For instance, AWS Rekognition’s "Violence" label includes subcategories such as "Physical Violence" and "Explosions and Blasts." We classify an image as NSFW if any subcategory score exceeds 0.5.
As hateful and harmful memes are an emerging challenge in moderation, commercial tools offer limited support in this area. We therefore include two SOTA academic methods: Pro-Cap~\cite{cao2023pro} and ExplainHM~\cite{lin2024towards}. Pro-Cap provides pre-trained models on both the Facebook Hateful Memes and Harmeme datasets, while ExplainHM is trained on Harmeme. We denote these as \texttt{Pro-Cap\_F}, \texttt{Pro-Cap\_H}, and \texttt{ExplainHM\_H}.
Together, these baselines represent leading methods across multiple NSFW categories and provide a strong foundation for evaluating \system{}'s effectiveness and generalizability.

To evaluate the robustness of \system{} across different language model backbones, we experiment with a diverse set of vision-language and large language models. We include three advanced vision-language models MiniCPM-V~\cite{yao2024minicpm}, DeepSeek-VL~\cite{lu2024DeepSeekvl}, and GPT-4o~\cite{openai2024gpt4o} along with the widely used LLaVA~\cite{liu2023llava}, covering both cutting-edge and established architectures, as well as open-source and proprietary systems. We also incorporate Llama-3-70B~\cite{meta2024llama3} to support advanced semantic reasoning and output aggregation.
All models were chosen based on their relevance and availability at the time of study. In the following sections, we analyze \system{}'s performance across these backbones on the previously introduced datasets, demonstrating its generalizability and robustness in real-world moderation scenarios.

\subsubsection{Prompt Settings}
\label{sec:settings}
We use the image categories and annotation criteria from each dataset to define rule inputs in our system. For common NSFW categories such as pornography, violence, and blood, we directly use the category names as rules. For domain-specific cases like school bully and hateful memes, we construct rule inputs based on the fine-grained definitions provided in the corresponding dataset papers.
Each rule is defined as a prohibition on the presence of the specified NSFW category in the image. This design choice ensures consistency with the dataset’s labeling standards, thereby enabling a fair and aligned evaluation. Moreover, the rule definitions are consistent with the content moderation policies of major social media platforms, further reinforcing their practical relevance. 

\noindent\textbf{Pornographic.}
The image type in system prompt $P_S$ is \textit{characters or scene}, as all pornographic images contain characters, whereas neutral images may include either characters or just a scene. The rule applied in $P_S$ is \textit{Ensure the image does not contain any pornographic materials.}

\noindent\textbf{Violence protest.}
The image type we input in system prompt $P_S$ is \textit{scene} because protest violence moderation requires the VLM to pay closer attention to the overall state of the scene.
The rule we input in $P_S$ is \textit{Ensure the image does not contain violence or depict violent acts with the intent to shock or disgust viewers.
}

\noindent\textbf{School bully.}
The image type we input in system prompt $P_S$ is \textit{school scene} because school violence moderation requires the VLM to pay closer attention to the overall state of the people in the school scene.
Based on the dataset description and the definition of school bully made by the U.S. government \cite{StopBullying_WhatIsBullying}, the rule we input in $P_S$ is \textit{Images must not contain any depictions of violence, bullying, physical altercations, or any aggressive behavior in a school setting or involving students.}

\noindent\textbf{Blood.}
The image type we input in system prompt $P_S$ is \textit{characters or scene} because the dataset includes human bleeding or gory and blood scene.
The rule in $P_S$ is \textit{Ensure the image does not contain any blood, gore, or depict bloody acts.}

\noindent\textbf{Hateful/harmful meme.}
The image type specified in the prompt $P_S$ is \textit{meme}.
Given the similarity in the definitions of harmful/hateful content across both datasets, we adopted Facebook's definition.
The rule we input in $P_S$ for the two meme dataset is: \textit{Images must not contain a direct or indirect attack on people based on characteristics, including ethnicity, race, nationality, immigration status, religion, caste, sex, gender identity, sexual orientation, and disability or disease.}


\subsubsection{Evaluation Metrics}
We report accuracy (Acc), precision (Pre), recall (Rec), and F1-score (F1) to comprehensively evaluate the moderation performance across each dataset.
For datasets that lack positive samples, specifically the blood and school bully datasets, we report only accuracy.

\subsection{Adaptability and Performance to Category-based Rules}
\label{sec:multicate}





\begin{table}[]
\centering
\caption{Comparing \system{} with the baselines on pornographic, school bully, and blood datasets.}
\resizebox{\linewidth}{!}{
\begin{tabular}{@{}ccccccc@{}}

\toprule
                       & \multicolumn{4}{c}{\textbf{Pornographic}} & \textbf{School Bully}     & \textbf{Blood}            \\ \cmidrule(l){2-5} \cmidrule(l){6-6}\cmidrule(l){7-7}
\textbf{Method}                 & \textbf{Acc}    & \textbf{Pre}    & \textbf{Rec}    & \textbf{F1}    & \textbf{Acc}              & \textbf{Acc}              \\ \midrule
Sightengine            & 0.919  & 0.919  & 0.973  & 0.945 & 0.407            & 0.834            \\
Imagaa                 & 0.829  & 0.670  & 0.984  & 0.797 & \textbackslash{} & \textbackslash{} \\
Amazon AWS Rekognition & 0.944  & 0.901  & 0.986  & 0.941 & 0.267            & 0.760            \\
Microsoft Azure        & 0.952  & 0.934  & 0.969  & 0.952 & 0.060            & 0.641            \\
Yahoo\_NSFW            & 0.930  & 0.918  & 0.940  & 0.929 & \textbackslash{} & \textbackslash{} \\ \hline
Ours\_MiniCPM-V        & 0.962  & 0.932  & \cellcolor{lightgray}0.997  & 0.964 & 0.780            &\cellcolor{lightgray}0.967            \\
Ours\_Deepseek-VL      & 0.909  & 0.864  & 0.970   & 0.914 & 0.733            & 0.926            \\
Ours\_LLaVA            & 0.902  & 0.923  & 0.877  & 0.900 & 0.870             & 0.952            \\
Ours\_GPT4o            &\cellcolor{lightgray} 0.985  &\cellcolor{lightgray}0.967  & 0.967  &\cellcolor{lightgray}0.967 & \cellcolor{lightgray}0.950             & 0.884            \\ \bottomrule
\end{tabular}
}
\label{tab:porntest}
\end{table}

\begin{table}[htbp]
\centering
\caption{Comparing \system{} with the baselines on two meme datasets. 
}
\label{tab:meme}
\resizebox{\linewidth}{!}{%
\begin{tabular}{@{}lllllllll@{}}
\toprule
\multicolumn{1}{c}{\multirow{2}{*}{\textbf{Method}}}           & \multicolumn{4}{c}{\textbf{Facebook Hateful unseen dataset}} & \multicolumn{4}{c}{\textbf{Harmeme COVID-19 dataset}} \\ \cmidrule(l){2-5} \cmidrule(l) {6-9}
                  & \textbf{Acc}          & \textbf{Pre}   & \textbf{Rec}          & \textbf{F1}           & \textbf{Acc}       & \textbf{Pre}       & \textbf{Rec}       & \textbf{F1}       \\ \midrule
ExplainHM\_H      & 0.610        & 0.472        & 0.145        & 0.222        & 0.783         & 0.673        & 0.735        & 0.703        \\
Pro-Cap\_F        & 0.699        & 0.741        & 0.301        & 0.429        & 0.614         & 0.314        & 0.087        & 0.135        \\
Pro-Cap\_H        & 0.611        & 0.364        & 0.050        & 0.091        & 0.777         & 0.636        & 0.849        & 0.727        \\ \hline
Ours\_MiniCPM-V   &  \cellcolor{lightgray}0.697 & 0.846 & 0.629 &  \cellcolor{lightgray}0.721 & 0.670  &  \cellcolor{lightgray}0.831 & 0.616 & 0.708 \\
Ours\_DeepSeek-VL & 0.636 & 0.714 &  \cellcolor{lightgray}0.697 & 0.705 & 0.687 & 0.762 & 0.751 & 0.756 \\
Ours\_LLaVA       & 0.627 & 0.726 & 0.647 & 0.684& 0.663  & 0.789 & 0.654 & 0.715 \\
Ours\_GPT4o       & 0.688 &  \cellcolor{lightgray}0.893 & 0.569& 0.695 &  \cellcolor{lightgray}0.744  & 0.792 &  \cellcolor{lightgray}0.820 &  \cellcolor{lightgray}0.805 \\ \bottomrule
\end{tabular}
}
\end{table}


In this section, we evaluate \system{}'s ability to adapt to moderation rules derived from distinct NSFW content categories and compare its performance with representative baseline methods.
Evaluation results for the pornographic and blood datasets are summarized in Table~\ref{tab:porntest},  and those for hateful/harmful memes are shown in Table~\ref{tab:meme}.
In these tables, rows labeled as "Ours\_X" represent the results of our system using different vision-language model backbones. In this section, we focus on the overall performance across NSFW categories, while a detailed discussion on the impact of different model backbones is provided in Section~\ref{sec:vlm}.
Overall, the results demonstrate that \system{} consistently achieves strong performance in all NSFW categories evaluated, confirming its adaptability to various moderation tasks. 

Furthermore, the results on pornographic and blood datasets demonstrate the effectiveness of \system{} in detecting image details. Specifically, \system{} consistently achieves the highest accuracy, precision, recall, and F1 scores, with top results exceeding 0.96. While baseline methods also perform relatively well, their success is due to the explicit visual objects, such as the sensitive body parts or visible blood, which simplify feature extraction. In contrast, by analyzing the images that existing methods fail to detect, we observe that \system{} outperforms all baselines, showcasing its ability to identify more subtle and localized NSFW content, such as small bloodstains or sensitive regions that may not be immediately noticeable. The example is shown in Appendix~\ref{app:details}.


The results on hateful/harmful datasets highlight \system{}'s ability to moderate complex and advanced semantic images. Detecting hateful or harmful content is particularly challenging due to the richness and complexity of advanced semantics, including rhetorical nuances and the combination of text and images. As a result, the detection accuracy for these categories is lower compared to other NSFW categories.
The results show that when the training set contains themes similar to those in the test set, the two comparative methods outperform our best result. However, when detecting different themes, their performance significantly decreases compared to ours. This indicates that their methods are highly dependent on the dataset and have poor generalization across different image content.
In contrast, our method operates in a training-free setting, and the metrics across various content types remain consistent without significant fluctuations. Notably, even under these conditions, our best performance is comparable to the performance of their models when trained on the same themes.

We further discuss the contribution of each design in capturing image details and understanding advanced semantics, along with related cases, in Section~\ref{sec:Ablation_exp}.


\begin{table*}[htbp]
\centering
\caption{Comparing \system{} with the baseline methods on violent protest datasets. The table shows the detection ratios of violent images at different levels of severity.}
\resizebox{0.85\linewidth}{!}{%
\begin{tabular}{llllllllllll}
\toprule
\diagbox{\textbf{Method}}{\textbf{Severity}}   & \textbf{0}  & \textbf{0-0.09} & \textbf{0.1-0.19} & \textbf{0.2-0.29} & \textbf{0.3-0.39} & \textbf{0.4-0.49} & \textbf{0.5-0.59} & \textbf{0.6-0.69} & \textbf{0.7-0.79} & \textbf{0.8-0.89} & \textbf{0.90-1} \\ \midrule
Sightengine            & 0           & 0               & 0                 & 0                 & 0                 & 0.020             & 0.049             & 0.297             & 0.230             & 0.470             & 0.571           \\
Amazon AWS Rekognition & 0.038       & 0               & 0.010             & 0.019             & 0.019             & 0.060             & 0.208             & 0.538             & 0.415             & 0.294             & 0.375           \\
Microsoft Azure        & 0.028       & 0               & 0               & 0               & 0               & 0.080              & 0.148             & 0.341             & 0.230             & 0.235             & 0.500             \\ \hline
Ours\_MiniCPM-V       & 0.133 & 0.067           & 0.181             & 0.107             & 0.186             & 0.190              & 0.534             & 0.868             & 0.953             & \cellcolor{lightgray}1                 & \cellcolor{lightgray}1               \\
Ours\_DeepSeek-VL      & 0.200         & 0.200             & 0.080              & 0.108             & 0.118             & 0.120             & 0.287             & 0.813             & 0.861             & 0.882             & \cellcolor{lightgray}1               \\
Ours\_LLaVA            & 0.304       & 0               & 0.255             & 0.323             & 0.421             & 0.460              & 0.643             & 0.857             & 0.861             & \cellcolor{lightgray}1                 & \cellcolor{lightgray}1               \\
Ours\_GPT4o            & 0.080        & 0               & 0.121             & 0.096             & 0.132             & 0.180              & 0.258             & \cellcolor{lightgray}0.930             & \cellcolor{lightgray}0.971             & \cellcolor{lightgray}1                 & \cellcolor{lightgray}1               \\ \bottomrule
\end{tabular}
\label{tab:protestacc}
}
\end{table*}

\subsection{Adaptability and Performance to Scenario-based Rules}
\label{sec:adaptivetorule}

\begin{table}[htbp]
\centering
\caption{Comparing \system{} with the baselines on porn datasets.}
\label{tab:cross_check}
\resizebox{\linewidth}{!}{%
\begin{tabular}{lccc}
\toprule
\multicolumn{1}{c}{\textbf{}}       & \multicolumn{2}{c}{\textbf{\begin{tabular}[c]{@{}c@{}}Protest dataset in \\ School Bully prompt\end{tabular}}} & \textbf{\begin{tabular}[c]{@{}c@{}}School Bully dataset in \\ Protest prompt\end{tabular}} \\ \cmidrule(l) {2-3} \cmidrule(l) {4-4}
\multicolumn{1}{c}{\textbf{Method}} & \textbf{Acc}                                      & \textbf{Not NSFW (0.6-0.7)}                                      & \textbf{Acc}   \\  \midrule
Ours\_MiniCPM-V            & 0.890                                          & 0.550                                                & 0.363                                                                             \\
Ours\_DeepSeek-VL          & 0.873                                          & 0.478                                                & 0.558                                                                             \\
Ours\_LLaVA                & 0.906                                          & 0.529                                                & 0.383                                                                             \\
Ours\_GPT4o                & 0.961                                          & 0.714                                                & 0.678                                                                             \\ \bottomrule
\end{tabular}
}
\end{table}

In this section, we evaluate \system{}'s adaptability to scenario-based moderation rules, where similar content may be subject to different enforcement standards depending on the application context. To assess this, we examine two cases: protest violence and school bullying. While both involve violent images, school bullying requires stricter moderation due to the presence of minors and child protection policies. Results for the violent protest dataset are shown in Table~\ref{tab:protestacc}, and results for school bullying are presented in Table~\ref{tab:porntest}.

Violent protest results show a positive correlation between detection accuracy and the original levels of violence in the images. 
Commercial tools struggle with moderating violence in specific contexts. 
Notably, for images with a violence level greater than 0.6, our approach achieved a detection accuracy exceeding 0.8, whereas the commercial tools only reached around 0.5.
This may be due to the complexity of the scene, which makes the subcategory less distinguishable or out of definition, as well as an insufficient understanding of the overall semantics of the image.
It confirms our concern that existing models trained solely on predefined datasets and labels may lack adaptability to real-world regulatory needs. 

Similarly, existing tools struggle with detecting school bully images, achieving a maximum accuracy of 0.407. 
Many images in the dataset, captured from surveillance perspectives, do not always position subjects within the ROI, complicating the capture of detailed interactions. 
Our method is able to capture these detail actions and character states for moderation, proving its effectiveness in reviewing details to address Challenge \uppercase\expandafter{\romannumeral2}. The example is shown in Appendix~\ref{app:details}.
Moreover, compared to generally defined violent behaviors, school bully tends to be less severe, which may cause its intensity to fall below the detection thresholds of existing methods.
Despite this, our method achieves high detection accuracy, with top results reaching 0.950, underscoring its robust generalization across diverse scenes and rules.
This also shows that our method remains effective even when moderation policies shift or specific groups are targeted in the review process.

We then designed an experiment to quantify the impact of prompt granularity on detection performance. By applying the fine‑grained “School Bully” rule in $P_s$ to the general violence test set and vice versa. We assessed each prompt’s behavior when used outside its target domain. 
Since the school bully dataset contains no positive samples, we perform cross-testing using only violent samples. Based on Table~\ref{tab:protestacc}, images with a violence score of 0.6 or higher are considered violent, and these images serve as a test set.

The results are in Table~\ref{tab:cross_check}. 
The first column shows moderation results using the protest violence dataset under the school bully rule. 
The school bully dataset is tested under the violence rule, which includes the entire set of data from the school bully dataset.
This cross-prompt setting illustrates how the rule in prompt granularity impacts performance outside its intended domain. The fine-grained "School Bully" rule achieves high precision on its target set but suffers a significant drop in recall and overall accuracy on broader violence data. In contrast, the general "Violence" rule maintains more balanced performance across both datasets.
When applying the "School Bully" rule to protest data, some violent images were missed mainly because (1) their violence scores (0.6–0.7) fell below the detection threshold (reported as “Not NSFW (0.6–0.7)” in the table), and (2) scenes like arson did not match school-related contexts. Conversely, using the "Violence" prompt on bully data allowed detection of several physical conflict scenes, showing that general prompts still capture relevant violations (see Table~\ref{tab:porntest}).
These quantitative insights inform our prompt‑engineering and deployment strategies: use specialized prompts when a narrow, high‑precision focus is required; use general prompts when coverage across diverse subtypes and the category is common and widely recognized; and decide whether to register multiple prompts and dynamically select or ensemble them at inference time. Ultimately, this experiment demonstrates that our framework not only reacts sensitively to different rule variants but also quantifies its specificity and generality in both matched and mismatched scenarios.



\subsection{Robustness Across Backbone Models}
\label{sec:vlm}
As a general framework, \system{} demonstrates consistently strong moderation performance across four distinct vision-language model backbones.
The rows labeled as “Ours\_X” in Table~\ref{tab:porntest}, Table~\ref{tab:protestacc}, and Table~\ref{tab:meme} present the results of our method applied to each of the four base models.
Although these VLMs differ significantly in terms of training strategies and model architectures, and exhibit notable performance variance in traditional VQA tasks\cite{yao2024minicpm}, the performance gap between VLMs observed under our framework in moderation tasks is small.
Specifically, the standard deviation of accuracy across the four VLMs is 0.035 for the pornographic dataset, followed by 0.083 for the school bully dataset, 0.031 for the blood dataset, 0.031 for the Facebook hateful dataset, and 0.032 for Harmeme dataset.
And the highest standard deviation is 0.051 for the violent protest dataset.
This demonstrates that our method is minimally affected by the choice of base VLM, highlighting its strong generalizability.


Furthermore, on datasets with explicit visual features, such as pornographic and blood content, \system{} achieves high accuracy across all backbones, particularly with MiniCPM-V and GPT-4o, which outperform others likely due to their superior visual recognition capabilities. Even with lower-performing models like DeepSeek-VL and LLaVA, \system{} maintains stable results, indicating strong overall robustness. This consistent performance is largely attributed to the explicit visual cues (e.g., sensitive body parts or blood), which are easier for vision-language models to detect.

Notably, \system{}'s strength lies in its framework design rather than prompt-level optimization. Unlike prompt tuning, which often requires task or model-specific adjustments~\cite{wang2023multitask}, we use a unified prompt format across tasks and backbones, focused only on task goals and reasoning. Its consistent performance across models highlights \system{}'s generality and practical effectiveness without complex prompt engineering.


\

\begin{table*}[htbp]
\centering

\caption{The results of using vanilla VLMs for NSFW moderation.}
\label{tab:directly_query}
\resizebox{1\linewidth}{!}{%
\begin{tabular}{lccccccccccccccc}
\toprule
\multicolumn{1}{c}{} & \multicolumn{5}{c}{\textbf{Pornographic}}        & \multicolumn{5}{c}{\textbf{Harmeme COVID-19}}    & \multicolumn{5}{c}{\textbf{Facebook Hateful unseen}} \\ \cmidrule(l){2-6}\cmidrule(l){7-11}  \cmidrule(l){12-16}
\multicolumn{1}{c}{\textbf{Method}} & \textbf{Acc}   & \textbf{Pre}   & \textbf{Rec}   & \textbf{F1}    & \textbf{Refusal} & \textbf{Acc}   & \textbf{Pre}   & \textbf{Rec}   & \textbf{F1}    & \textbf{Refusal} & \textbf{Acc}    & \textbf{Pre}    & \textbf{Rec}    & \textbf{F1}     & \textbf{Refusal} \\ \midrule
Deepseek-vl-7b  & 0.492(-0.417) & 0.499        & 0.966        & 0.658       & 0.708            & 0.440(-0.247)  & 0.657        & 0.077        & 0.138       & 0.003            & 0.471(-0.165) & 0.800          & 0.205        & 0.327       & 0.001            \\
MiniCPM-V-2.5   & 0.490(-0.472) & 0.497        & 0.778        & 0.607       & 0.772            & 0.438(-0.232) & 0.532        & 0.279        & 0.366       & 0.03             & 0.539(-0.158) & 0.692        & 0.472        & 0.561       & 0.009            \\
LLaVA-v1.6      & 0.896(-0.006) & 0.841        & 0.979        & 0.905       & 0.007            & 0.503(-0.160)  & 0.643        & 0.326        & 0.433       & 0.184            & 0.562(-0.065) & 0.691        & 0.54         & 0.606       & 0.079            \\
GPT4o           & 0.927(-0.058) & 0.95         & 0.919        & 0.934       & 0                & 0.658(-0.086) & 0.686        & 0.777        & 0.728       & 0            & 0.631(-0.057)  & 0.756        & 0.592        & 0.664       & 0.002      
      
      \\ \bottomrule
\end{tabular}
}

\end{table*}

\vspace{-5mm}

\section{Further Analysis}
In addition to the core evaluations, we conduct a series of in-depth experiments to analyze the limitations of vanilla VLMs, assess the contribution of our designs, evaluate the reliability of existing benchmark datasets, and examine the practicality of \system{} in realistic moderation scenarios, aiming to provide deeper insights into the challenges and dynamics of moderation tasks.
\textbf{First}, we test vanilla VLMs to demonstrate that existing VLMs alone are inadequate for effective NSFW moderation.
\textbf{Second}, we conduct a step-wise evaluation of our linear review pipeline, assessing the incremental performance gains introduced by each review stage. This analysis quantifies the contribution of each component to address Challenge \uppercase\expandafter{\romannumeral2} and Challenge \uppercase\expandafter{\romannumeral3} and the overall decision quality.
\textbf{Third}, in contrast to classification-based methods, our system produces not only final NSFW judgments but also image descriptions and reasoning traces. When analyzing prediction-label mismatches, we \textit{surprisingly} discover that a non-negligible portion of these discrepancies are due to inconsistent label samples. This finding reveals the system’s potential for assisting data quality control and auditing.
\textbf{Finally}, we evaluate \system{} under real-world conditions using in-the-wild data. These extended experiments provide deeper insights into its robustness and practical effectiveness.

\subsection{Comparison with Vanilla VLMs}

We followed \cite{qu2024unsafebench} to directly query four different VLMs and evaluate their performance in NSFW content moderation. The evaluation was conducted on three datasets with clearly defined positive and negative samples. The detection metrics are shown in Table~\ref{tab:directly_query}. Notably, the "Refusal" column indicates the proportion of incorrect predictions caused by the model's refusal to respond or make a judgment.
Values in parentheses indicate the differences from our moderation accuracy.

The results show that VLMs indeed possess potential for NSFW moderation, particularly in identifying common NSFW categories. However, their accuracy is not yet sufficient for practical deployment. This highlights the necessity of designing a highly adaptable moderation framework.
Moreover, due to safety alignment mechanisms within large models, there is a high probability of refusal to moderate, especially in the pornography category. This aligns with the findings reported in \cite{qu2024unsafebench}. Among the evaluated models, GPT4o demonstrates notably strong performance. Since our prompts do not require the model to generate NSFW content directly, GPT4o typically does not refuse to respond, even if it flags the presence of NSFW elements in the image.
We even found that, in some cases, the provided justifications were inaccurate. Detailed analysis can be found in Appendix~\ref{app:unreliable}.
Additionally, the results show that the performance of Vanilla VLMs varies significantly. However, after applying our framework, even models with weaker performance demonstrate strong moderation capabilities, validating the robustness of our approach across different backbone models mentioned in Section~\ref{sec:vlm}.

\subsection{Design Contribution Evaluation}
\label{sec:Ablation_exp}

\begin{figure*}[htbp]
  \centering
   \includegraphics[width=1\linewidth]{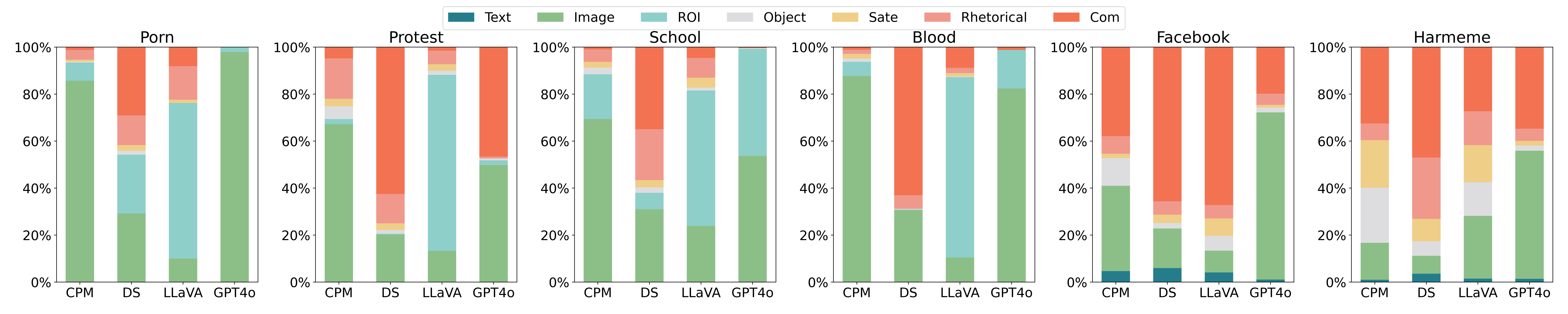}
   \caption{Contributions of each module to the moderation of NSFW datasets}
   \label{fig:rq2}
\end{figure*}
We analyze the contributions of the standard image moderation and advanced semantic analysis modules in \system{}. Both modules operate progressively and utilize multiple review prompts to identify malicious images at various stages. By analyzing the proportion of NSFW images detected at each step, we highlight their importance. Notably, every NSFW image identified at a later stage but missed earlier underscores the unique contribution of that step to the overall process.

In the standard image moderation, we designed three steps with three progressive review prompts: text content moderation (Text), overall image moderation (Image), and adaptive region of interest zoom-in (ROI). 
In the advanced semantics analysis, we designed four steps with four description-based progressive review prompts: object description moderation (Object), state description moderation (State), rhetorical device description (Rhetorical), and comprehensive moderation (Com).
In cases such as porn, violent protest (Protest), school violence (School), and blood datasets (Blood), where there is little to no effective text, text content moderation was not utilized. For meme moderation (Facebook and Harmeme), where the overall meaning and the integration of image and text are more critical, we opted not to use the adaptive region of interest zoom-in, prioritizing efficiency instead.
Figure \ref{fig:rq2} presents the contributions of these seven moderation steps in detecting NSFW samples under four different base VLMs when detecting six datasets.
From the bottom to the top of the bar chart, following the moderation process, it illustrates the proportion detected by each module when identifying negative samples.
This experiment demonstrates three key points.
First, each step in the process enhances NSFW image detection by identifying content that previous steps may have missed to varying degrees. 
This indicates that some NSFW features overlooked by previous modules are addressed by the subsequent moderation modules.
Second, advanced semantic analysis is particularly effective in conducting high-level semantic moderation. In the two datasets of memes, advanced semantics analysis contributed between 27.8\% and 88.8\% to the moderation process. 
This underscores the necessity and effectiveness of our design and confirms that VLMs indeed struggle to understand advanced image semantics under normal circumstances.
These findings highlight the importance of a multi-step approach and the value of incorporating advanced semantic analysis in achieving comprehensive and accurate image moderation.

\subsection{Label Noise Discovery and Analysis}
\label{sec:manual}

Compared to existing classifiers, \system{} outputs final results with image descriptions and reasoning. By analyzing these outputs alongside the tested images, we identified some inconsistent or controversial labels in the dataset. 
To this end, we implemented a manual annotation method to verify the labels. To ensure accuracy and minimize bias, the annotation strictly followed Google's annotation guidelines \cite{GoogleCloud_DataLabeling}. 
The details of the annotation can be found in Appendix \ref{app:annotation}.
Annotators conducted a manual review of the experiment results for the pornographic, violent protest, blood, and hateful/harmful memes datasets.
For the pornographic, violent protest, and blood datasets, due to ethical considerations, the manual review was carried out exclusively by three authors, similar to the approach in \cite{Guo2024ModeratingIO}. No additional personnel were exposed to these sensitive image types. For the hateful/harmful meme dataset, 80 annotators with verified expertise in security and content safety were employed to annotate the results.

For the pornographic dataset, annotators manually reviewed 16 incorrectly classified positive samples and identified six as implicit pornographic images. Further analysis of incorrectly classified negative samples revealed that some normal images were inconsistent as negative porn samples. This inconsistency occurred because, while the samples were derived from pornographic videos, not all frames contained explicit content, leading to misclassification.
To better understand the dataset's inconsistency issues, we utilized miniCPM-V as the base VLM and tested all 8,595 images in the porn test set.
Among the 82 images that were incorrectly classified as normal, a manual review confirmed that 62 of them were indeed normal or, at the very least, should not have been classified as pornographic.
Examples are provided in Appendix \ref{app:C}.

For the violent protest dataset, one notable observation is that our system was able to detect some violent images even at the 0 level (the column indicating a severity of 0 in Table~\ref{tab:protestacc}). These images, while not depicting violent actions or states, contain aggressive text or symbols related to racial discrimination, leading \system{} to classify them as violent. However, it is important to note that the dataset's annotations focus exclusively on explicit violent actions and states, such as assault, vandalism, looting, and arson. Examples of these samples are provided in Appendix \ref{app:D}.

For the blood dataset, using the results of MiniCPM-V as a baseline, annotators identified 43 images that were incorrectly classified as normal. A manual annotation confirmed that 30 of these images were indeed normal or, at the very least, should not have been classified as blood images.
We observed that the misclassified content often involved red pigments or horror-themed visuals that did not depict actual bloody scenes. Such similarities in appearance can introduce biases during data collection. Examples are provided in Appendix \ref{app:E}.

For hateful/harmful meme datasets, the complex meanings conveyed by these images make them more prone to controversy.
To gain a better understanding, we tested all 2,000 unseen Facebook Hateful images and all 3,544 Harmeme COVID-19 images based on the four VLMs.
Then, annotators identified potentially incorrect samples by considering those where the moderation results from at least three VLMs were inconsistent with the original labels. This process revealed 489 such images in the Facebook Hateful unseen dataset and 683 in the Harmeme COVID-19 dataset. These flagged samples were subjected to manual annotation.
The results of manual annotation are shown in Appendix~\ref{app:manual_meme}
There are numerous factors that can lead to inconsistency, including the use of discriminatory language. This may be due to the original annotators' limited understanding of derogatory terms for specific groups, as well as the evolution of word meanings in response to societal changes.
Examples of the wrongly labeled images are provided in Appendix \ref{app:F}.

We have reported the observed label inconsistencies to the respective dataset publishers, and some of them have been partially confirmed.
According to the dataset publishers, there are two main reasons for these inconsistencies. First, the data was collected through keyword matching on search engines or social media platforms, which may have resulted in some noisy samples that were not properly filtered. Second, due to evolving regulatory standards and societal developments, the criteria for judging certain samples may differ between the time of collection and the present.
This highlights the practical challenges of NSFW content moderation: existing methods are prone to noisy samples in the training data, and the trained models struggle to adapt to changes in regulatory frameworks and societal values. In contrast, our moderation framework does not rely on a training dataset and can flexibly adjust moderation tasks based on evolving policies, underscoring both the necessity and urgency of this research.

In conclusion, our method offers significant advantages: as a training-free approach, it prevents performance degradation caused by incorrect labels or data poisoning during training. Its ability to provide intermediate analysis and reasoning helps identify and rectify inconsistent label samples, reducing biases in downstream tasks, particularly in harmful content detection, where such biases could lead to discrimination against specific groups. Additionally, our method reduces the costs of manual data collection and annotation while alleviating the psychological distress associated with labeling harmful content.

\subsection{NSFW Moderation in the Wild}
\label{sec:realworld_exp}
To demonstrate that our method can adapt to moderating topics related to ongoing events, we collected a set of meme images based on current trending events for evaluation.
We collected 817 memes related to recent trending and sensitive topics from the well-known website Memedroid \cite{memedroid_rules}.
Memedroid is a platform focused on internet humor and image sharing, featuring mainly funny pictures and memes. 
As of the most recent data, Memedroid has over 10 million users worldwide. 
Users can browse, share, and comment on various types of humorous images and memes on the platform.
Memedroid has its own moderation rules and review policies \cite{memedroid_rules}.
We conducted searches using sensitive keywords such as "Russia," "Israel," "Ukraine," "Jews," "Choke," "Freaky," "Kinky," "Sexual," "Dank," 
and "Dirty Mind," which include both trending topics (Russia-Ukraine, Israel-Palestine) and sexually suggestive terms. 
Moreover, these images were published between May and September 2024.
Compared to the data in the published dataset, these trending topics reflect current events, making them time-sensitive.

The experiment settings are the same as hateful/harmful meme moderation. We selected MiniCPM-V and GPT4o as the base VLM. In the aforementioned experiments, these two models demonstrated superior performance.


We had annotators manually label the images, following the annotation method described in Section \ref{sec:manual}.
For the total of 817 images, \system{} using MiniCPM-V detected 416 images as containing violations and 419 such images based on GPT4o. 
Both models agreed that 337 of these images were in violation. 
We confirmed that 259 of these memes indeed contained hateful content. 
However, among these images, ProCap\_H could only detect 51, ProCap\_F detected only 13, and ExplainHM\_H could detect only 37. 
This indicates that in real-world scenarios, our method can identify images that existing methods fail to detect. 
This demonstrates the practical effectiveness of our approach and highlights that, compared to existing methods, our method is better suited for moderating a broader range of image content. 
Examples of detected harmful samples are provided in Appendix \ref{app:G}.

\section{Discussion}

\noindent\textbf{Limitation.} 
Our study has several limitations. First, the performance is still dependent on the VLM's inherent capabilities. While our workflow enhances VLM accuracy, less capable models struggle with advanced semantics, leading to gaps in understanding human societal norms. This issue partly explains our suboptimal performance in meme experiments. 
Additionally, our approach remains susceptible to adversarial examples or jailbreaking attacks, which are tied to the VLM's intrinsic resilience and robustness. Moreover, the base VLM's inference speed may affect the moderation efficiency.
Second, our study currently focuses on pornographic, violent, blood, hateful, and harmful content. 
Other types of NSFW content exist, including content related to religion. 
Expanding our analysis to more NSFW categories could enhance our understanding of online content safety challenges.


\noindent\textbf{Future work.}
As we continue to enhance our approach to NSFW content moderation, several avenues for improvement have emerged.
First, fine-tuning the VLM specifically for image moderation scenarios is advisable. In NSFW domains where moderation is urgently needed, collecting and manually annotating samples for fine-tuning can help the VLM develop a more objective and fair understanding of images in line with specific regulations. This approach can also address issues like the model's reluctance to provide responses.
Second, NSFW content moderation should not be confined to images alone. It is crucial to develop more universal moderation methods that can be applied to text, video, audio, and other multimedia content. Additionally, moderation approaches should be capable of multimodal integration, enabling comprehensive moderation across various types of media.

\section{Conclusion}

We introduce \system{}, a general NSFW moderation framework. We introduce VLMs for a multimodal alignment strategy that adapts to diverse NSFW regulations and scenarios.
To address challenges in capturing image details, we implement techniques including visual optimization and adaptive zoom-in.
We also introduce iterative output aggregation to tackle issues with VLMs refusing responses or giving sanitized answers. 
\system{} includes image moderation process including standard image moderation, and advanced semantic analysis.
Extensive experiments show that \system{} outperforms significantly existing methods and our method
exhibits strong adaptability across categories, scenarios, and
base VLMs. 
We also identify inconsistent samples and detecting images missed by other approaches.

\section{Ethics Consideration}\label{app:ethic}
We carefully address the potential ethical issues in this work. Our study was reviewed and approved by the author's institution's supervisory department (SD) and Institutional Review Board (IRB). Additionally, we implemented the following steps.


\textbf{Content warning}. This paper contains content that may be offensive or uncomfortable for some readers. To mitigate this, we have censored the most distressing parts by applying mosaics to the sensitive sections and have included a warning on the first page of the paper.

\textbf{Well-being support.} This work involves annotating NSFW images, which may expose annotators to disturbing content and affect their psychological well-being. To ensure ethical practices and provide comprehensive support, we established strict annotation guidelines and conducted ethical reviews before, during, and after the process. Annotators may skip or stop at any time and are limited to labeling no more than 50 images per dataset to safeguard mental health and ensure fairness. This task is entirely voluntary, and healthcare support, including optional psychological counseling, is offered if needed. Annotators are compensated based on completed work, and we take full responsibility for any harm incurred, contingent on hospital-issued documentation. To date, no psychological health concerns have been reported. We continue regular well-being check-ins and remain open to refining our protocols based on feedback.

\textbf{Dataset lifecycle.} We prioritize security and privacy throughout the dataset lifecycle. Similar to prior work \cite{Guo2024ModeratingIO, paudel2024enabling}, we exclusively used publicly available datasets for all harmful content, with proper permissions obtained from the platforms and publishers. Personal information was removed from all examples in this study.
All harmful content used in this study comes from publicly available datasets obtained with proper permissions, including porn and Facebook Hateful Memes from Kaggle, the bully dataset from Roboflow, the Harmeme dataset from its official GitHub repository \cite{pramanick2021momenta}, and the protest violence and blood datasets with authors’ consent \cite{won2017protest,wang2023tovilag}. Meme images were collected from Memedroid in accordance with its terms of use, without further editing or additional harmful data collection. All experiments were conducted in a controlled environment compliant with local laws and regulations, with strict controls on data inflow and outflow to prevent any dissemination of harmful content. For relabeled data, we will only release image filenames and corrected labels, excluding harmful material. Due to IRB policy, images from Memedroid will not be released. All examples in the paper have been sanitized.

\textbf{Tool abuse.} From the defenders' perspective, we propose a general moderation framework with the core objective of moderating harmful content to prevent its dissemination. The use of harmful data is solely for performance evaluation to enable more effective regulation. Additionally, our method can improve moderation efficiency and reduce the burden of manual moderation, protecting human well-being from exposure to harmful content.
From the defenders' perspective, we propose a general moderation framework aimed at curbing the dissemination of harmful content. Harmful data is used solely for performance evaluation to enhance regulatory effectiveness. Our method improves moderation efficiency and reduces reliance on manual review, thereby protecting individuals from unnecessary exposure to harmful material.
While the framework serves the public interest by enhancing moderation capabilities, it is equally important to consider its ethical implications and potential for misuse. In particular, although designed to promote societal well-being in line with the Menlo Report’s principles of beneficence and respect for persons \cite{kenneally2012menlo}, \system{} could be exploited by malicious actors. For example, biased platform providers may misuse it to unfairly regulate or suppress specific users or content, leading to censorship. To address this risk, behavioral and traffic analysis tools may help detect biased moderation patterns \cite{zhu2013velocity,athanasopoulos2011censmon,jones2014automated}. We also advocate for independent third-party oversight of platform rule-setting and enforcement to ensure transparency, accountability, and safeguard against abuse.

\newpage
\bibliographystyle{IEEEtran}
\bibliography{moderation}

\begin{thebibliography}{10}
\providecommand{\url}[1]{#1}
\csname url@samestyle\endcsname
\providecommand{\newblock}{\relax}
\providecommand{\bibinfo}[2]{#2}
\providecommand{\BIBentrySTDinterwordspacing}{\spaceskip=0pt\relax}
\providecommand{\BIBentryALTinterwordstretchfactor}{4}
\providecommand{\BIBentryALTinterwordspacing}{\spaceskip=\fontdimen2\font plus
\BIBentryALTinterwordstretchfactor\fontdimen3\font minus \fontdimen4\font\relax}
\providecommand{\BIBforeignlanguage}[2]{{%
\expandafter\ifx\csname l@#1\endcsname\relax
\typeout{** WARNING: IEEEtran.bst: No hyphenation pattern has been}%
\typeout{** loaded for the language `#1'. Using the pattern for}%
\typeout{** the default language instead.}%
\else
\language=\csname l@#1\endcsname
\fi
#2}}
\providecommand{\BIBdecl}{\relax}
\BIBdecl

\bibitem{NSFW}
\BIBentryALTinterwordspacing
Wikipedia, ``Not safe for work,'' \url{https://en.wikipedia.org/wiki/Not_safe_for_work}, 2024, accessed: 2024-07-25. [Online]. Available: \url{https://en.wikipedia.org/wiki/Not_safe_for_work}
\BIBentrySTDinterwordspacing

\bibitem{momo_challenge_hoax}
``Momo challenge hoax,'' \url{https://en.wikipedia.org/wiki/Momo_Challenge_hoax}, April 2025, wikipedia, the free encyclopedia.

\bibitem{nguyen2020multi}
Q.-H. Nguyen, H.-L. Tran, T.-T. Nguyen, D.-D. Phan, D.-L. Vu \emph{et~al.}, ``Multi-level detector for pornographic content using cnn models,'' in \emph{2020 RIVF international conference on computing and communication technologies (RIVF)}.\hskip 1em plus 0.5em minus 0.4em\relax IEEE, 2020, pp. 1--5.

\bibitem{tabone2021pornographic}
A.~Tabone, K.~Camilleri, A.~Bonnici, S.~Cristina, R.~Farrugia, and M.~Borg, ``Pornographic content classification using deep-learning,'' in \emph{Proceedings of the 21st ACM symposium on document engineering}, 2021, pp. 1--10.

\bibitem{geremias2022motion}
J.~Geremias, E.~K. Viegas, A.~S. Britto~Jr, and A.~O. Santin, ``A motion-based approach for real-time detection of pornographic content in videos,'' in \emph{Proceedings of the 37th ACM/SIGAPP Symposium on Applied Computing}, 2022, pp. 1066--1073.

\bibitem{Guo2024ModeratingIO}
K.~Guo, A.~Utkarsh, W.~Ding, I.~Ondracek, Z.~Zhao, G.~Freeman, N.~Vishwamitra, and H.~Hu, ``Moderating illicit online image promotion for unsafe user generated content games using large vision language models,'' in \emph{33st {USENIX} Security Symposium, {USENIX} Security 2024, Philadelphia, PA, USA, August 14-16, 2024}.\hskip 1em plus 0.5em minus 0.4em\relax {USENIX} Association, 2024.

\bibitem{pang2022audiovisual}
W.~Pang, W.~Xie, Q.~He, Y.~Li, and J.~Yang, ``Audiovisual dependency attention for violence detection in videos,'' \emph{IEEE Transactions on Multimedia}, vol.~25, pp. 4922--4932, 2022.

\bibitem{ullah2021ai}
F.~U.~M. Ullah, K.~Muhammad, I.~U. Haq, N.~Khan, A.~A. Heidari, S.~W. Baik, and V.~H.~C. de~Albuquerque, ``Ai-assisted edge vision for violence detection in iot-based industrial surveillance networks,'' \emph{IEEE Transactions on Industrial Informatics}, vol.~18, no.~8, pp. 5359--5370, 2021.

\bibitem{liu2022decouple}
T.~Liu, C.~Zhang, K.-M. Lam, and J.~Kong, ``Decouple and resolve: transformer-based models for online anomaly detection from weakly labeled videos,'' \emph{IEEE Transactions on Information Forensics and Security}, vol.~18, pp. 15--28, 2022.

\bibitem{kiela2020hateful}
D.~Kiela, H.~Firooz, A.~Mohan, V.~Goswami, A.~Singh, P.~Ringshia, and D.~Testuggine, ``The hateful memes challenge: Detecting hate speech in multimodal memes,'' \emph{Advances in neural information processing systems}, vol.~33, pp. 2611--2624, 2020.

\bibitem{he2023you}
X.~He, S.~Zannettou, Y.~Shen, and Y.~Zhang, ``You only prompt once: On the capabilities of prompt learning on large language models to tackle toxic content,'' \emph{arXiv preprint arXiv:2308.05596}, 2023.

\bibitem{cao2023pro}
R.~Cao, M.~S. Hee, A.~Kuek, W.-H. Chong, R.~K.-W. Lee, and J.~Jiang, ``Pro-cap: Leveraging a frozen vision-language model for hateful meme detection,'' in \emph{Proceedings of the 31st ACM International Conference on Multimedia}, 2023, pp. 5244--5252.

\bibitem{pramanick2021momenta}
S.~Pramanick, S.~Sharma, D.~Dimitrov, M.~S. Akhtar, P.~Nakov, and T.~Chakraborty, ``Momenta: A multimodal framework for detecting harmful memes and their targets,'' in \emph{Findings of the Association for Computational Linguistics: EMNLP 2021}, 2021, pp. 4439--4455.

\bibitem{sightengine}
{Sightengine}, ``Image moderation,'' \url{https://sightengine.com/image-moderation}, 2024, accessed: 2024-08-17.

\bibitem{clarifai}
Clarifai, ``Content moderation,'' \url{https://www.clarifai.com/solutions/content-moderation}, 2024, accessed: 2024-08-17.

\bibitem{aws_rekognition}
{Amazon Web Services}, ``Amazon rekognition,'' \url{https://aws.amazon.com/cn/rekognition/}, accessed: 2024-08-17.

\bibitem{azure_ai_content_safety}
{Microsoft Azure}, ``Ai content safety,'' \url{https://azure.microsoft.com/en-us/products/ai-services/ai-content-safety/}, 2024, accessed: 2024-08-17.

\bibitem{aldahoul2024advancing}
N.~AlDahoul, M.~J.~T. Tan, H.~R. Kasireddy, and Y.~Zaki, ``Advancing content moderation: Evaluating large language models for detecting sensitive content across text, images, and videos,'' \emph{arXiv preprint arXiv:2411.17123}, 2024.

\bibitem{teenagers_exposion}
T.~Cohen, ``Teenagers exposed to 'horrific' content online - and this survey reveals the scale of the problem,'' \url{https://news.sky.com/story/teenagers-exposed-to-horrific-content-online-and-this-survey-reveals-the-scale-of-the-problem-13331556}, 2025, sky News.

\bibitem{riccio2024exposed}
P.~Riccio, T.~Hofmann, and N.~Oliver, ``Exposed or erased: Algorithmic censorship of nudity in art,'' in \emph{Proceedings of the CHI Conference on Human Factors in Computing Systems}, 2024, pp. 1--17.

\bibitem{oliveira2021detection}
M.~Oliveira~Franca, ``Detection and categorization of suggestive thumbnails: A step towards a safer internet,'' 2021.

\bibitem{leu2024auditing}
W.~Leu, Y.~Nakashima, and N.~Garcia, ``Auditing image-based nsfw classifiers for content filtering,'' in \emph{The 2024 ACM Conference on Fairness, Accountability, and Transparency}, 2024, pp. 1163--1173.

\bibitem{pilipets2022nipples}
E.~Pilipets and S.~Paasonen, ``Nipples, memes, and algorithmic failure: Nsfw critique of tumblr censorship,'' \emph{New Media \& Society}, vol.~24, no.~6, pp. 1459--1480, 2022.

\bibitem{sarridis2022leveraging}
I.~Sarridis, C.~Koutlis, O.~Papadopoulou, and S.~Papadopoulos, ``Leveraging large-scale multimedia datasets to refine content moderation models,'' in \emph{2022 IEEE Eighth International Conference on Multimedia Big Data (BigMM)}.\hskip 1em plus 0.5em minus 0.4em\relax IEEE, 2022, pp. 125--132.

\bibitem{hong2024s}
R.~Hong, W.~Agnew, T.~Kohno, and J.~Morgenstern, ``Who's in and who's out? a case study of multimodal clip-filtering in datacomp,'' \emph{arXiv preprint arXiv:2405.08209}, 2024.

\bibitem{qu2024unsafebench}
Y.~Qu, X.~Shen, Y.~Wu, M.~Backes, S.~Zannettou, and Y.~Zhang, ``Unsafebench: Benchmarking image safety classifiers on real-world and ai-generated images,'' \emph{arXiv preprint arXiv:2405.03486}, 2024.

\bibitem{evaluation_software}
N.~Aldahoul, H.~Abdul~Karim, M.~A. Momo, M.~Sy, and M.~J. Tan, ``Evaluation of content moderation software for nudity and pornography detection in various scenarios,'' 07 2023.

\bibitem{yuan2019stealthy}
K.~Yuan, D.~Tang, X.~Liao, X.~Wang, X.~Feng, Y.~Chen, M.~Sun, H.~Lu, and K.~Zhang, ``Stealthy porn: Understanding real-world adversarial images for illicit online promotion,'' in \emph{2019 IEEE Symposium on Security and Privacy (SP)}.\hskip 1em plus 0.5em minus 0.4em\relax IEEE, 2019, pp. 952--966.

\bibitem{karabulut2023automatic}
D.~Karabulut, C.~Ozcinar, and G.~Anbarjafari, ``Automatic content moderation on social media,'' \emph{Multimedia Tools and Applications}, vol.~82, no.~3, pp. 4439--4463, 2023.

\bibitem{Meta_AdultNuditySexualActivity}
I.~Meta~Platforms, ``Adult nudity and sexual activity,'' \url{https://transparency.meta.com/en-us/policies/community-standards/adult-nudity-sexual-activity/}, 2024, accessed: 2024-12-26.

\bibitem{qumeme}
Y.~M. X. S.~Y. Qu, N.~Y.~M. Backes, and S.~Z.~Y. Zhang, ``From meme to threat: On the hateful meme understanding and induced hateful content generation in open-source vision language models.''

\bibitem{wang2023tovilag}
X.~Wang, X.~Yi, H.~Jiang, S.~Zhou, Z.~Wei, and X.~Xie, ``Tovilag: Your visual-language generative model is also an evildoer,'' in \emph{The 2023 Conference on Empirical Methods in Natural Language Processing}, 2024.

\bibitem{laranjeira2022seeing}
C.~Laranjeira~da Silva, J.~Macedo, S.~Avila, and J.~dos Santos, ``Seeing without looking: Analysis pipeline for child sexual abuse datasets,'' in \emph{Proceedings of the 2022 ACM Conference on Fairness, Accountability, and Transparency}, 2022, pp. 2189--2205.

\bibitem{devlin2019bert}
J.~Devlin, M.-W. Chang, K.~Lee, and K.~Toutanova, ``Bert: Pre-training of deep bidirectional transformers for language understanding. arxiv,'' \emph{arXiv preprint arXiv:1810.04805}, 2019.

\bibitem{lin2024towards}
H.~Lin, Z.~Luo, W.~Gao, J.~Ma, B.~Wang, and R.~Yang, ``Towards explainable harmful meme detection through multimodal debate between large language models,'' in \emph{Proceedings of the ACM on Web Conference 2024}, 2024, pp. 2359--2370.

\bibitem{tabassum2024investigating}
M.~Tabassum, A.~Mackey, A.~Schuett, and A.~Lerner, ``Investigating moderation challenges to combating hate and harassment: The case of mod-admin power dynamics and feature misuse on reddit,'' in \emph{30th USENIX Security Symposium (USENIX Security 24). USENIX Association}, 2024.

\bibitem{arunasalam2024understanding}
A.~Arunasalam, H.~Farrukh, E.~Tekcan, and Z.~B. Celik, ``Understanding the security and privacy implications of online toxic content on refugees,'' in \emph{USENIX Security Symposium}, 2024.

\bibitem{chu2022behind}
A.~Chu, A.~Arunasalam, M.~O. Ozmen, and Z.~B. Celik, ``Behind the tube: Exploitative monetization of content on $\{$YouTube$\}$,'' in \emph{31st USENIX Security Symposium (USENIX Security 22)}, 2022, pp. 2171--2188.

\bibitem{vu2023no}
A.~V. Vu, A.~Hutchings, and R.~Anderson, ``No easy way out: the effectiveness of deplatforming an extremist forum to suppress hate and harassment,'' \emph{arXiv preprint arXiv:2304.07037}, 2023.

\bibitem{imagga}
{Imagga}, ``Adult content moderation,'' \url{https://imagga.com/solutions/adult-content-moderation}, 2024, accessed: 2024-08-17.

\bibitem{radford2021learning}
A.~Radford, J.~W. Kim, C.~Hallacy, A.~Ramesh, G.~Goh, S.~Agarwal, G.~Sastry, A.~Askell, P.~Mishkin, J.~Clark \emph{et~al.}, ``Learning transferable visual models from natural language supervision,'' in \emph{International conference on machine learning}.\hskip 1em plus 0.5em minus 0.4em\relax PMLR, 2021, pp. 8748--8763.

\bibitem{li2022blip}
J.~Li, D.~Li, C.~Xiong, and S.~Hoi, ``Blip: Bootstrapping language-image pre-training for unified vision-language understanding and generation,'' in \emph{International conference on machine learning}.\hskip 1em plus 0.5em minus 0.4em\relax PMLR, 2022, pp. 12\,888--12\,900.

\bibitem{liu2023llava}
H.~Liu, C.~Li, Q.~Wu, and Y.~J. Lee, ``Visual instruction tuning,'' 2023.

\bibitem{lu2024DeepSeekvl}
H.~Lu, W.~Liu, B.~Zhang, B.~Wang, K.~Dong, B.~Liu, J.~Sun, T.~Ren, Z.~Li, H.~Yang, Y.~Sun, C.~Deng, H.~Xu, Z.~Xie, and C.~Ruan, ``Deepseek-vl: Towards real-world vision-language understanding,'' 2024.

\bibitem{yao2024minicpm}
Y.~Yao, T.~Yu, A.~Zhang, C.~Wang, J.~Cui, H.~Zhu, T.~Cai, H.~Li, W.~Zhao, Z.~He \emph{et~al.}, ``Minicpm-v: A gpt-4v level mllm on your phone,'' \emph{arXiv preprint arXiv:2408.01800}, 2024.

\bibitem{openai2024gpt4o}
\BIBentryALTinterwordspacing
{OpenAI}, ``Gpt-4o system card,'' 2024, accessed: 2024-08-20. [Online]. Available: \url{https://openai.com/index/gpt-4o-system-card/}
\BIBentrySTDinterwordspacing

\bibitem{chen2024lion}
G.~Chen, L.~Shen, R.~Shao, X.~Deng, and L.~Nie, ``Lion: Empowering multimodal large language model with dual-level visual knowledge,'' in \emph{Proceedings of the IEEE/CVF Conference on Computer Vision and Pattern Recognition}, 2024, pp. 26\,540--26\,550.

\bibitem{yuan2024osprey}
Y.~Yuan, W.~Li, J.~Liu, D.~Tang, X.~Luo, C.~Qin, L.~Zhang, and J.~Zhu, ``Osprey: Pixel understanding with visual instruction tuning,'' in \emph{Proceedings of the IEEE/CVF Conference on Computer Vision and Pattern Recognition}, 2024, pp. 28\,202--28\,211.

\bibitem{wu2024v}
P.~Wu and S.~Xie, ``V*: Guided visual search as a core mechanism in multimodal llms,'' in \emph{Proceedings of the IEEE/CVF Conference on Computer Vision and Pattern Recognition}, 2024, pp. 13\,084--13\,094.

\bibitem{zhang2023gpt4roi}
S.~Zhang, P.~Sun, S.~Chen, M.~Xiao, W.~Shao, W.~Zhang, Y.~Liu, K.~Chen, and P.~Luo, ``Gpt4roi: Instruction tuning large language model on region-of-interest,'' \emph{arXiv preprint arXiv:2307.03601}, 2023.

\bibitem{zhao2023chatspot}
L.~Zhao, E.~Yu, Z.~Ge, J.~Yang, H.~Wei, H.~Zhou, J.~Sun, Y.~Peng, R.~Dong, C.~Han \emph{et~al.}, ``Chatspot: Bootstrapping multimodal llms via precise referring instruction tuning,'' \emph{arXiv preprint arXiv:2307.09474}, 2023.

\bibitem{lu2023lyrics}
J.~Lu, R.~Gan, D.~Zhang, X.~Wu, Z.~Wu, R.~Sun, J.~Zhang, P.~Zhang, and Y.~Song, ``Lyrics: Boosting fine-grained language-vision alignment and comprehension via semantic-aware visual objects,'' \emph{arXiv preprint arXiv:2312.05278}, 2023.

\bibitem{wei2022chain}
J.~Wei, X.~Wang, D.~Schuurmans, M.~Bosma, F.~Xia, E.~Chi, Q.~V. Le, D.~Zhou \emph{et~al.}, ``Chain-of-thought prompting elicits reasoning in large language models,'' \emph{Advances in neural information processing systems}, vol.~35, pp. 24\,824--24\,837, 2022.

\bibitem{kojima2022large}
T.~Kojima, S.~S. Gu, M.~Reid, Y.~Matsuo, and Y.~Iwasawa, ``Large language models are zero-shot reasoners,'' \emph{Advances in neural information processing systems}, vol.~35, pp. 22\,199--22\,213, 2022.

\bibitem{yao2024tree}
S.~Yao, D.~Yu, J.~Zhao, I.~Shafran, T.~Griffiths, Y.~Cao, and K.~Narasimhan, ``Tree of thoughts: Deliberate problem solving with large language models,'' \emph{Advances in Neural Information Processing Systems}, vol.~36, 2024.

\bibitem{Meta_CommunityStandards}
I.~Meta~Platforms, ``Facebook community standards,'' \url{https://transparency.meta.com/policies/community-standards}, 2024, accessed: 2024-12-26.

\bibitem{YouTube_CommunityGuidelines}
\BIBentryALTinterwordspacing
YouTube, ``Community guidelines,'' 2024, accessed: 2024-12-26. [Online]. Available: \url{https://www.youtube.com/howyoutubeworks/policies/community-guidelines/}
\BIBentrySTDinterwordspacing

\bibitem{X_Rules}
X.~Corporation, ``X rules,'' \url{https://help.x.com/en/rules-and-policies/x-rules}, 2024, accessed: 2024-12-26.

\bibitem{farhan2024visual}
M.~Farhan~Ishmam, I.~Tashdeed, T.~Asir~Saadat, M.~Hamjajul~Ashmafee, R.~M. Kamal, D.~Abu, and M.~A. Hossain, ``Visual robustness benchmark for visual question answering (vqa),'' \emph{arXiv e-prints}, pp. arXiv--2407, 2024.

\bibitem{wang2021real}
X.~Wang, L.~Xie, C.~Dong, and Y.~Shan, ``Real-esrgan: Training real-world blind super-resolution with pure synthetic data,'' in \emph{Proceedings of the IEEE/CVF international conference on computer vision}, 2021, pp. 1905--1914.

\bibitem{huang2023open}
X.~Huang, Y.-J. Huang, Y.~Zhang, W.~Tian, R.~Feng, Y.~Zhang, Y.~Xie, Y.~Li, and L.~Zhang, ``Open-set image tagging with multi-grained text supervision,'' \emph{arXiv e-prints}, pp. arXiv--2310, 2023.

\bibitem{liu2023grounding}
S.~Liu, Z.~Zeng, T.~Ren, F.~Li, H.~Zhang, J.~Yang, C.~Li, J.~Yang, H.~Su, J.~Zhu \emph{et~al.}, ``Grounding dino: Marrying dino with grounded pre-training for open-set object detection,'' \emph{arXiv preprint arXiv:2303.05499}, 2023.

\bibitem{zhang2024large}
Z.~Zhang, G.~Shen, G.~Tao, S.~Cheng, and X.~Zhang, ``On large language models’ resilience to coercive interrogation,'' in \emph{2024 IEEE Symposium on Security and Privacy (SP)}.\hskip 1em plus 0.5em minus 0.4em\relax IEEE Computer Society, 2024, pp. 252--252.

\bibitem{renze2024effect}
M.~Renze and E.~Guven, ``The effect of sampling temperature on problem solving in large language models,'' \emph{arXiv preprint arXiv:2402.05201}, 2024.

\bibitem{pramanick2021detecting}
S.~Pramanick, D.~Dimitrov, R.~Mukherjee, S.~Sharma, M.~S. Akhtar, P.~Nakov, and T.~Chakraborty, ``Detecting harmful memes and their targets,'' in \emph{Findings of the Association for Computational Linguistics: ACL-IJCNLP 2021}, 2021, pp. 2783--2796.

\bibitem{won2017protest}
D.~Won, Z.~C. Steinert-Threlkeld, and J.~Joo, ``Protest activity detection and perceived violence estimation from social media images,'' in \emph{Proceedings of the 25th ACM international conference on Multimedia}, 2017, pp. 786--794.

\bibitem{old-violence-0tsy1_dataset}
\BIBentryALTinterwordspacing
H.~C.~M. university~of Technology, ``old violence dataset,'' \url{ https://universe.roboflow.com/ho-chi-minh-university-of-technology-eqlzy/old-violence-0tsy1 }, nov 2023, visited on 2024-08-20. [Online]. Available: \url{https://universe.roboflow.com/ho-chi-minh-university-of-technology-eqlzy/old-violence-0tsy1}
\BIBentrySTDinterwordspacing

\bibitem{YouTube_ViolentGraphicContent}
YouTube, ``Violent or graphic content,'' \url{https://support.google.com/youtube/answer/2802008}, 2024, accessed: 2024-12-26.

\bibitem{StopBullying_WhatIsBullying}
U.~D. of~Health and H.~Services, ``What is bullying?'' \url{https://www.stopbullying.gov/bullying/what-is-bullying}, 2024, accessed: 2024-12-26.

\bibitem{yahoo2016open_nsfw}
Y.~Inc., ``Yahoo open nsfw,'' \url{https://github.com/yahoo/open_nsfw}, 2016, accessed: 2024-08-25.

\bibitem{meta2024llama3}
\BIBentryALTinterwordspacing
Meta, ``Llama 3: Model cards and prompt formats,'' 2024, accessed: 2024-08-26. [Online]. Available: \url{https://llama.meta.com/docs/model-cards-and-prompt-formats/meta-llama-3}
\BIBentrySTDinterwordspacing

\bibitem{wang2023multitask}
Z.~Wang, R.~Panda, L.~Karlinsky, R.~Feris, H.~Sun, and Y.~Kim, ``Multitask prompt tuning enables parameter-efficient transfer learning,'' \emph{arXiv preprint arXiv:2303.02861}, 2023.

\bibitem{GoogleCloud_DataLabeling}
\BIBentryALTinterwordspacing
G.~Cloud. (2024) Data labeling. Accessed: 2024-12-26. [Online]. Available: \url{https://cloud.google.com/use-cases/data-labeling}
\BIBentrySTDinterwordspacing

\bibitem{memedroid_rules}
Memedroid, ``Memedroid rules,'' \url{https://www.memedroid.com/rules}, 2024, accessed: 2024-08-25.

\bibitem{paudel2024enabling}
P.~Paudel, M.~H. Saeed, R.~Auger, C.~Wells, and G.~Stringhini, ``Enabling contextual soft moderation on social media through contrastive textual deviation,'' in \emph{33rd USENIX Security Symposium (USENIX Security 24)}, 2024, pp. 4409--4426.

\bibitem{kenneally2012menlo}
E.~Kenneally and D.~Dittrich, ``The menlo report: Ethical principles guiding information and communication technology research,'' \emph{Available at SSRN 2445102}, 2012.

\bibitem{zhu2013velocity}
T.~Zhu, D.~Phipps, A.~Pridgen, J.~R. Crandall, and D.~S. Wallach, ``The velocity of censorship:$\{$High-Fidelity$\}$ detection of microblog post deletions,'' in \emph{22nd USENIX Security Symposium (USENIX Security 13)}, 2013, pp. 227--240.

\bibitem{athanasopoulos2011censmon}
E.~Athanasopoulos, S.~Ioannidis, and A.~Sfakianakis, ``$\{$CensMon$\}$: A web censorship monitor,'' in \emph{USENIX Workshop on Free and Open Communications on the Internet (FOCI 11)}, 2011.

\bibitem{jones2014automated}
B.~Jones, T.-W. Lee, N.~Feamster, and P.~Gill, ``Automated detection and fingerprinting of censorship block pages,'' in \emph{Proceedings of the 2014 Conference on Internet Measurement Conference}, 2014, pp. 299--304.

\bibitem{wikipedia2024covfefe}
\BIBentryALTinterwordspacing
Wikipedia, ``Covfefe,'' \url{https://en.wikipedia.org/wiki/Covfefe}, 2024, accessed: 2024-08-29. [Online]. Available: \url{https://en.wikipedia.org/wiki/Covfefe}
\BIBentrySTDinterwordspacing

\end{thebibliography}

\clearpage
\appendix

\subsection{Annotation Steps} \label{app:annotation}
The annotation mainly includes the following steps:
1) Define labeling guidelines: our annotation employs a binary approach, explicitly using $P_s$ as the rule. Samples that violate this rule are classified as negative, otherwise as positive.
Clear labeling guidelines can help annotators better understand the labeling tasks.
2) Select labeling tools: we set up an annotation server to facilitate access for annotators.
This helps annotators minimize external interference and facilitates the annotation process.
3) Train Annotators: We train and test annotators using samples where the results of our method are consistent with the original labels, ensuring that there is no overlap between the training set and the testing set. Annotators are only qualified to perform annotations after passing the test with an accuracy of 80\% or greater.
This reduces annotators' subjective judgments and annotation biases, and enhances their labeling capabilities.
4) Implement quality Control: The annotation results for each sample must be agreed upon by at least two annotators.

\subsection{The Prompt of Iterative Moderation Output Aggregation}\label{app:A}
The prompt of iterative moderation output aggregation $T_L$ for the LLM is:

\textit{You are provided with 10 descriptions evaluating a specific section of an image to determine if it violates the rule \{rule\}. Based on these descriptions, determine if the image violates the rule and provide a comprehensive justification for your conclusion. Exclude descriptions that refuse to provide an evaluation, give an ambiguous or incorrect answer, or discuss violations unrelated to the rule. 
    Descriptions: \{moderations\}
    }

Please provide the rule and the ten moderation descriptions from VLM to input in parentheses.

The prompt of iterative descripiton output aggregation $T_D$ for the LLM is:

{\itshape You are a professional image reviewer. Below are 10 descriptions of an image from different perspectives. Please create one cohesive and detailed summary description in one paragraph based on these descriptions. 
    When integrating the descriptions, follow these requirements:
    
    1. Focus primarily on objective descriptions of objects and characters.
    
    2. Include specific names of individuals, events, or backgrounds mentioned in any descriptions.
    
    3. Exclude any descriptions of atmosphere, sense, or suggestions related to objects, characters, or the overall image.
    
    4. If any description indicates a policy violation, state the violated policies directly instead of providing a summary description.
    
    5. Format your response as follows: Summary Description: [your description or policy violation]

    "Descriptions: \{descriptions\}
}

Please provide ten descriptions from VLM in parentheses.

\subsection{Comprehensive Moderation Prompt}\label{app:B}
The comprehensive prompt design $T_C$ for the final comprehensive evaluation is:

{\itshape
You are a professional image reviewer. You have received descriptions of an image based on three different perspectives. Your task is to analyze these descriptions and determine if the image complies with the following rule: \{rule\}

Here are the descriptions provided:
Image Text (text on the image):
\{image\_text\}
First Description (about the objects and people in the image):
\{object\_description\}
Second Description (about the objects and people state in the image):
\{state\_description\}
Third Description (about the image meaning):
\{shetorical\_ devices\_description\}

Based on the given descriptions, perform the following tasks:
1. Create a cohesive and detailed summary description in one paragraph, integrating the provided descriptions while excluding any subjective interpretations or suggestions.
2. Determine if the image violates the specified rule. If any description indicates a policy violation, state the violated policies directly instead of providing a summary description.
3. Ensure your answer follows the format: 
    Summary Description: [your description or policy violation]
    Compliance with Rule: [Yes/No]

Example Analysis:

Summary Description: A red car is parked by the sidewalk with John Doe standing next to it. The background includes a park with green trees. The car is positioned parallel to the sidewalk, and John Doe is leaning against it, looking at the park. The setting is a bright afternoon with people walking their dogs in the background.

Compliance with Rule: No

Your Task:
Based on the descriptions provided above, complete the analysis as demonstrated in the example.
}

Please provide the specific rules, image text $I_T$, objects description $R_{D1}$, state description $R_{D2}$ and rhetorical devices description $R_{D3}$ to input in parentheses. 

\begin{figure}[htbp]

      \begin{subfigure}[b]{0.23\textwidth}    
          \centering
          \includegraphics[width=1.7in,height=1.29in]{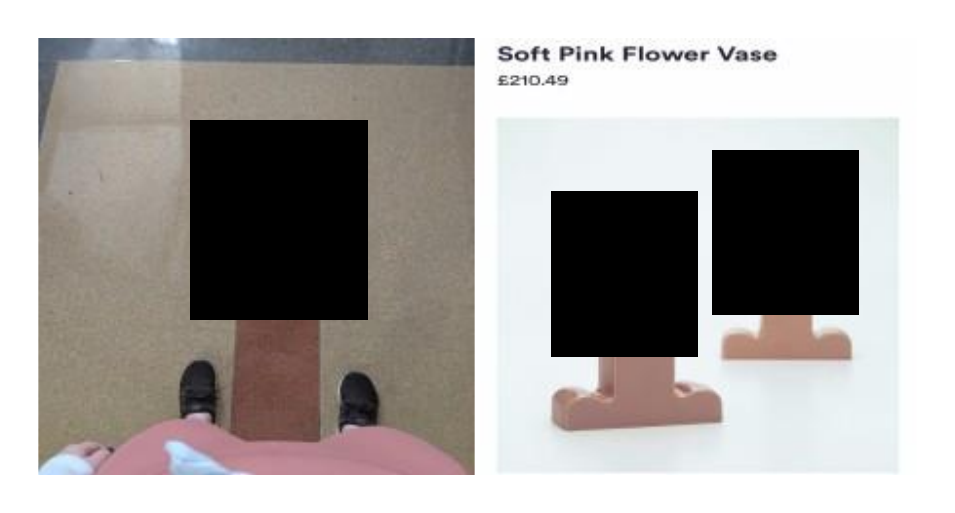}
          \caption{Samples of the incorrect original label as "neutral".}
          \label{fig:in_neutral}
      \end{subfigure}
      \hspace{0.1cm}
         \begin{subfigure}[b]{0.23\textwidth}
             \centering
             \includegraphics[width=1.7in,height=1.29in]{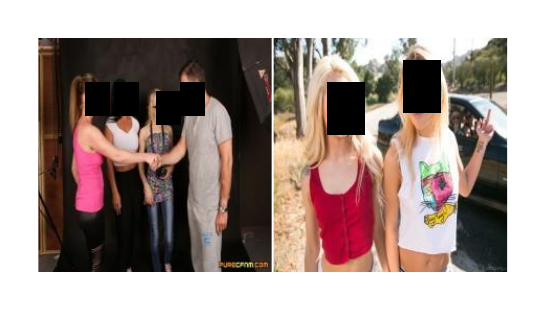}
             \caption{Samples of the incorrect original label as "porn".}
             \label{fig:in_porn}
         \end{subfigure}
             \caption{Samples of the incorrect original label in porn dataset.}  
             \label{fig:in_sample_porn}     
 \end{figure}

\begin{figure}[htbp]
  \centering
   \includegraphics[width=0.5\linewidth]{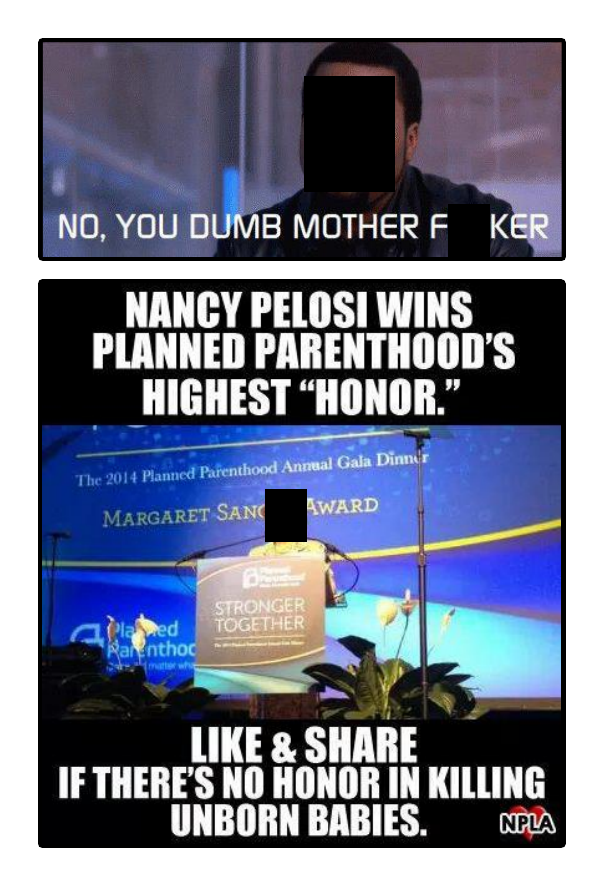}
   \caption{Content violent image for violence level 0 in the violent protest dataset.}
   \label{fig:in_protest}
\end{figure}

\subsection{Examples of Moderation in Detailed Regions}
\label{app:details}

\begin{figure}[htbp]
  \centering
   \includegraphics[width=1\linewidth]{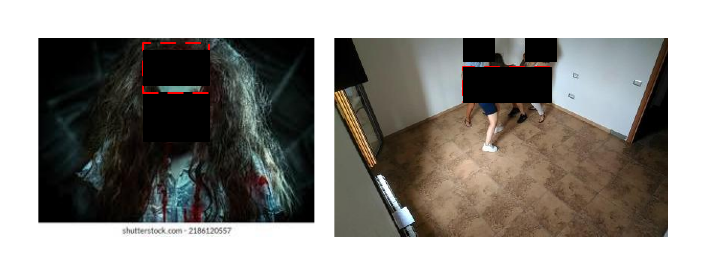}
   \caption{NSFW content in details of the images. The NSFW content is in the red box.}
   \label{fig:sample_details}
\end{figure}

Figure~\ref{fig:sample_details} shows samples that were not identified as NSFW by the comparison methods but were flagged as NSFW by our approach through the ROI zoom-in moderation. These two samples are from the blood and school bully datasets, respectively. In these samples, the NSFW region is not the main content of the image. For instance, in the image considered to be blood, but the core content of the original image is a terrifying character. There is some blood on its head. In the school bully image, the violent content is located at the edge of the image, with the overall image being from a surveillance perspective.

\subsection{VLM's Unreliable Justification.}
\label{app:unreliable}
We found that using vanilla VLMs under NSFW moderation task tends to generate unfounded associations when providing results in some cases. For instance, when directly evaluating violent samples, those with a violence score of 0-0.1 were often misclassified as violent. 
Specifically, Deepseek-VL classifies 0.816 of these samples as violent, MiniCPM-V classifies 0.576, and LLaVA classifies 0.456.
The models' justification was that normal protest scenes could easily be associated with conflict, implying the presence of violence. Such speculative moderation significantly undermines the reliability of the results.
This prevents us from conducting reliable evaluations and reporting valid metrics on these datasets.
Although this issue does not occur across all NSFW categories, due to the model's inherent reasoning capabilities, it poses a potential risk during moderation if no constraints are applied.

\subsection{Annotation Results in Hateful/Harmful Datasets}
\label{app:manual_meme}
The results of manual annotation in Hateful/Harmful datasets are shown in Table \ref{tab:manual_meme}.
The second row (Inconsistent) of the table shows the number of samples where the original label was found to be incorrect after manual annotation, while the third row (Consistent) shows the number of samples where the original label was confirmed to be correct.
\begin{table}[htbp]
\centering
\caption{Manual annotation (Man) results compared to Original label (Ori).}
\label{tab:manual_meme}
\resizebox{1\linewidth}{!}{%
\begin{tabular}{@{}lllllll@{}}
\toprule
      & \multicolumn{3}{l}{\textbf{Harmeme COVID-19 dataset}}           & \multicolumn{3}{l}{\textbf{Facebook Hateful unseen dataset}}                 \\ \cmidrule(l){2-4}\cmidrule(l){5-7} 
\diagbox{\textbf{Man}}{\textbf{Ori}}  & \multicolumn{1}{c}{\textbf{Total}} & \ \textbf{Not harmful} &   \textbf{Harmful} & \textbf{Total} &\textbf{Not hateful} & \textbf{Hateful} \\ \midrule
\textbf{Total }                                                        & 683                       & 371                                                             & 312                                                         & 489   & 382                                                             & 107                                                         \\
\begin{tabular}[c]{@{}l@{}}\textbf{Inconsistent}\end{tabular} & 525                       & 258                                                             & 267                                                         & 356   & 293                                                             & 63                                                          \\
\begin{tabular}[c]{@{}l@{}}\textbf{Consistent}\end{tabular}   & 158                       & 113                                                             & 45                                                          & 133   & 89                                                              & 44                                                          \\ \bottomrule
\end{tabular}

}
\end{table}

\subsection{Examples of Incorrect Labels in the Porn Dataset}
\label{app:C}
Figure \ref{fig:in_sample_porn} illustrates samples from the porn dataset where the original labels were incorrect. In these cases, the original label as "neutral" was applied to the images in Figure \ref{fig:in_neutral}. The labeling error occurred because these images symbolically represent sexual organs using normal content, which can easily lead to a dirty mind. The original label as "porn" was applied to the images in Figure \ref{fig:in_porn}. The labeling error here stems from the fact that, when viewed in isolation, the content of these images does not contain explicit pornographic material.

\subsection{Examples of Incorrect Labels in the Violent Protest  Dataset}
\label{app:D}

Figure \ref{fig:in_protest} shows images from the violent protest dataset that have a violence level of 0 but contain violent language. 
These images do not depict violent actions, as the dataset's violence level annotation primarily focused on physical violence and did not account for text. 
In contrast, \system{} provides a more comprehensive review process, enabling it to detect violent content conveyed through text.

\subsection{Examples of Incorrect Labels in the Blood Dataset}
\label{app:E}

\begin{figure}
  \centering
   \includegraphics[width=1\linewidth]{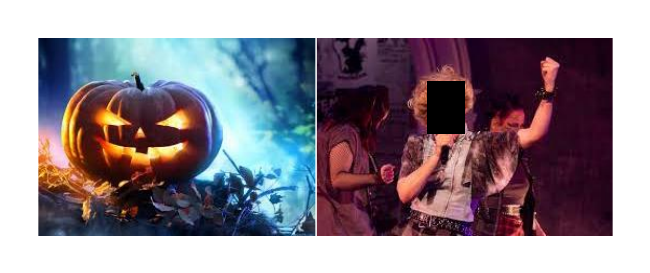}
   \caption{No blood image in blood dataset.}
   \label{fig:in_blood}
\end{figure}

Figure \ref{fig:in_blood} illustrates samples from the blood dataset that are not blood images.
Although the images feature red or some frightening colors, they do not actually contain blood content.

\subsection{Examples of Incorrect Labels in the Meme Datasets}
\label{app:F}

\begin{figure}[t]

      \begin{subfigure}[b]{0.23\textwidth}    
          \centering
          \includegraphics[width=1.7in,height=2.4in]{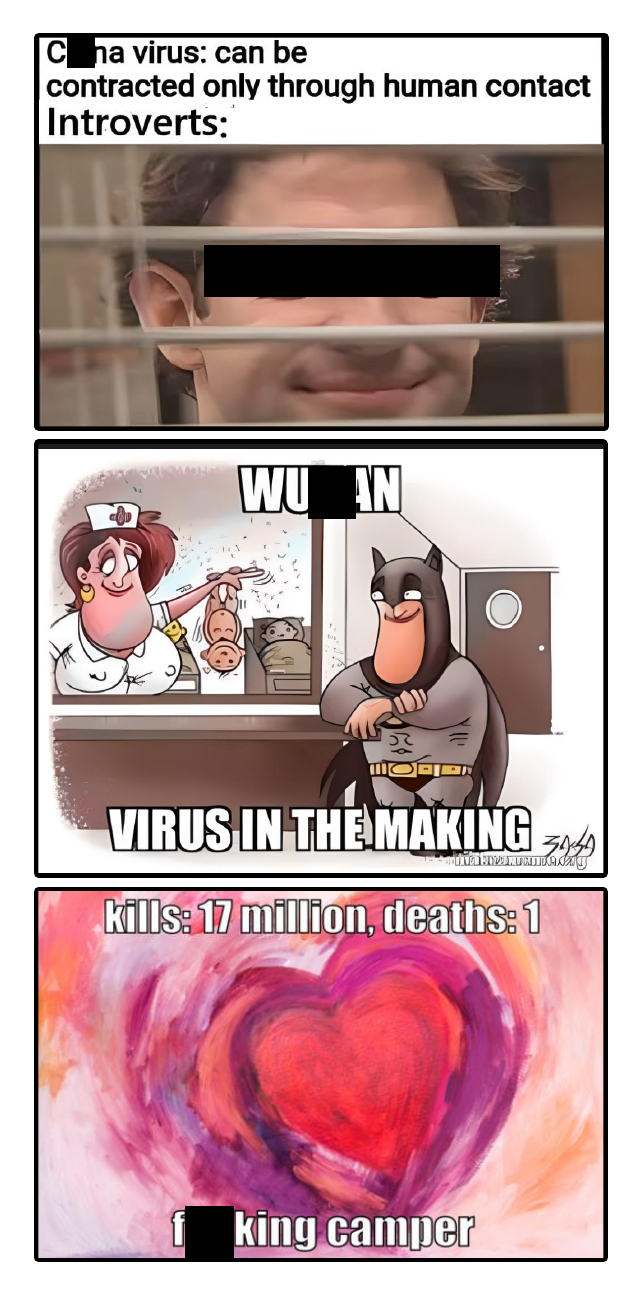}
          \caption{Samples of the original label as "not harmful/hateful".}
          \label{fig:in_right}
      \end{subfigure}
      \hspace{0.1cm}
         \begin{subfigure}[b]{0.23\textwidth}
             \centering
             \includegraphics[width=1.7in,height=2.4in]{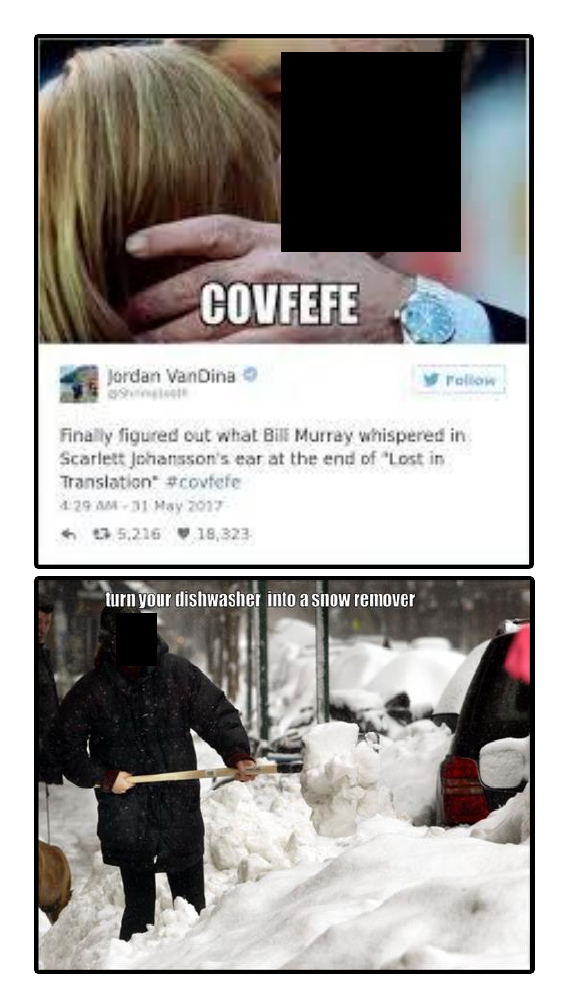}
             \caption{Samples of the original label as "harmful/hateful".}
             \label{fig:in_wrong}
         \end{subfigure}
             \caption{Samples of the incorrect original label in two meme datasets.}  
             \label{fig:in_meme}     
 \end{figure}

Figure \ref{fig:in_meme} illustrates samples from the two harmful meme datasets where the original labels were incorrect. In these cases, the original label as "not harmful/hateful" was applied to the images in Figure \ref{fig:in_right}. The labeling error occurred because these images contain discriminatory or aggressive content targeting specific groups or regions. 
The original label as "harmful/hateful" was applied to the images in Figure \ref{fig:in_wrong}. 
The content in these images is not directed at any specific individual or group.
For example, "COVFEFE" is merely a playful reference to a typo made by Donald Trump, which emerged in 2017 and has no connection to COVID-19. In fact, Trump himself has even joked about the term \cite{wikipedia2024covfefe}.

\begin{figure}[htbp]

      \begin{subfigure}[b]{0.23\textwidth}    
          \centering
          \includegraphics[width=1.7in,height=2.4in]{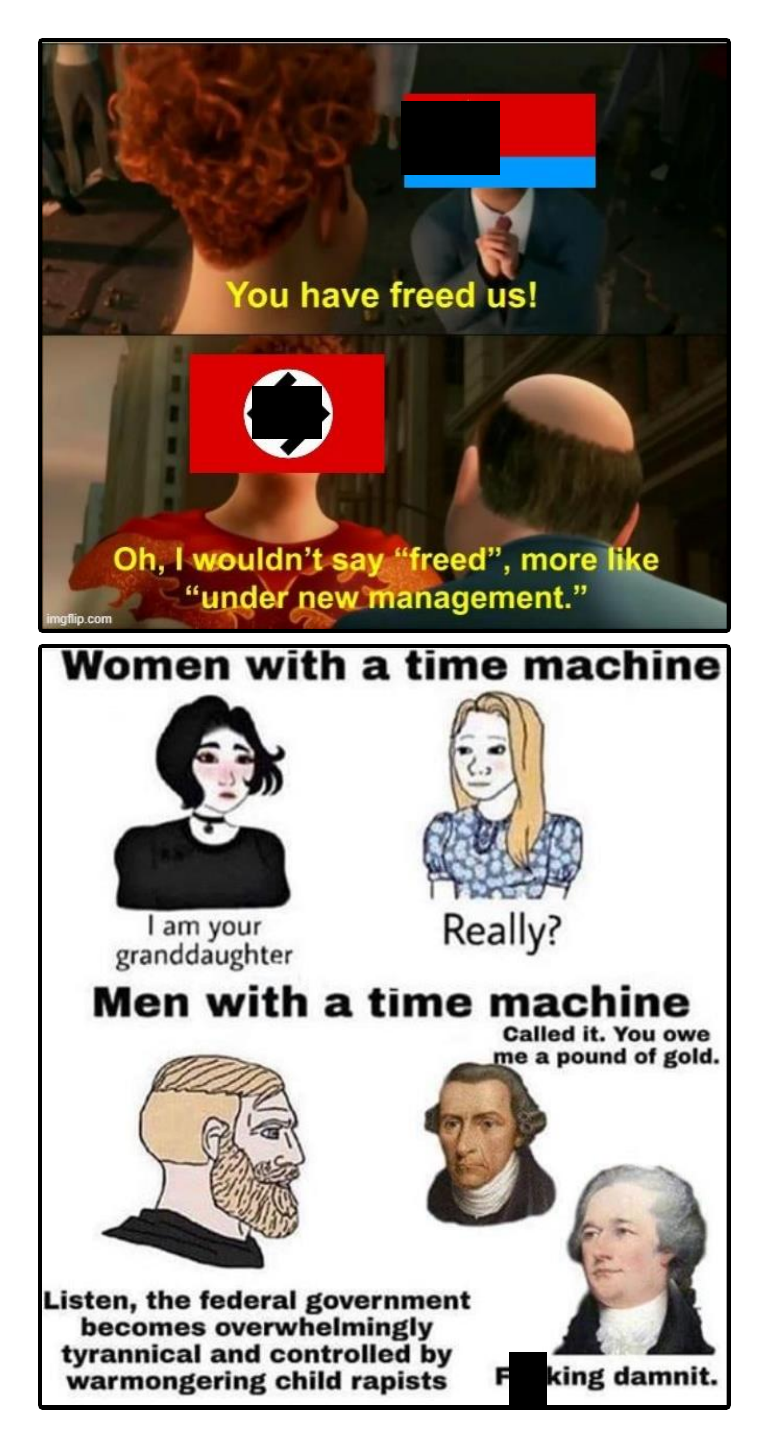}
          \caption{Samples of the harmful meme image detected by \system{} and the comparison methods.}
          \label{fig:real_common}
      \end{subfigure}
      \hspace{0.1cm}
         \begin{subfigure}[b]{0.23\textwidth}
             \centering
             \includegraphics[width=1.7in,height=2.4in]{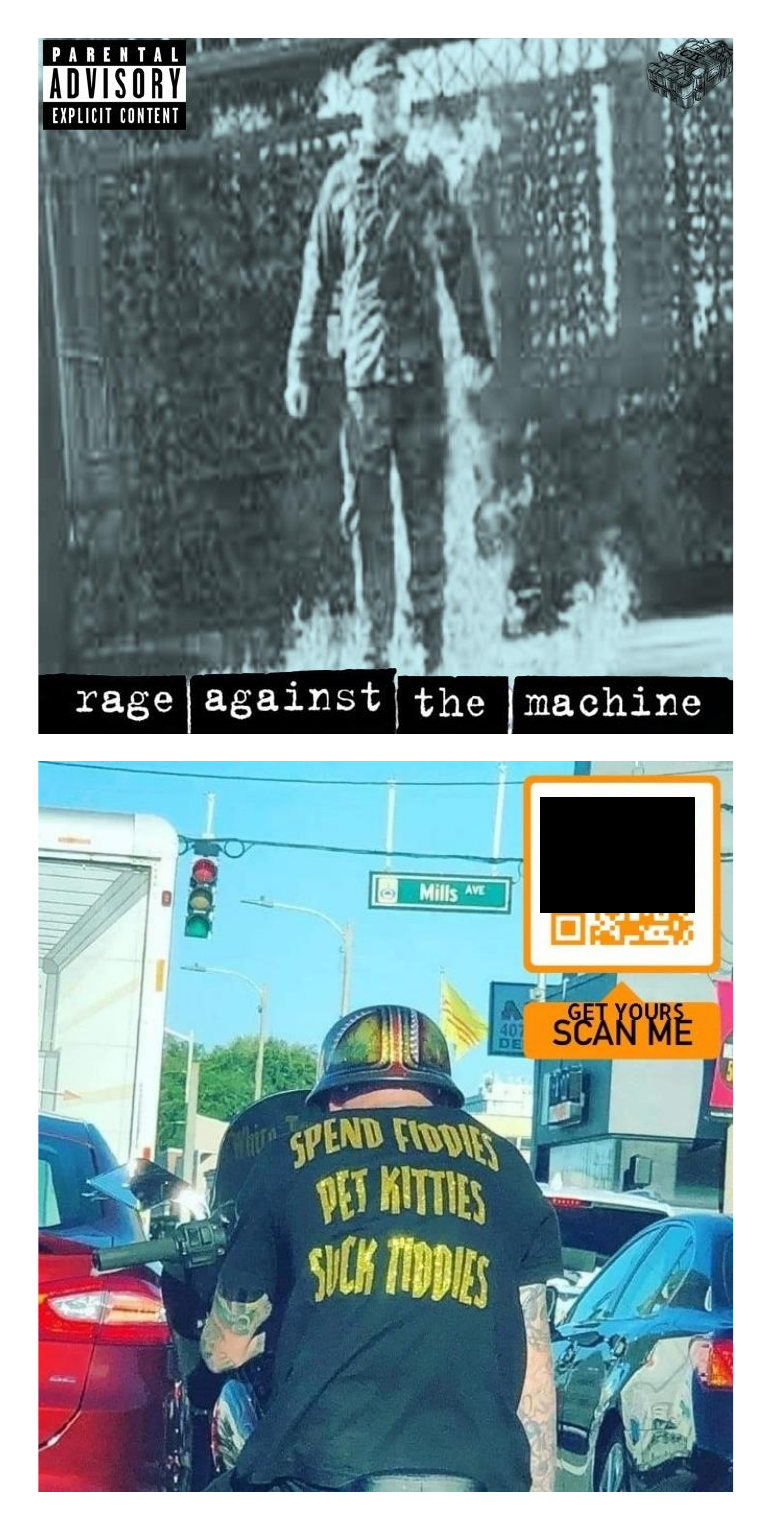}
             \caption{Samples of the harmful meme image detected only by \system{}.}
             \label{fig:real_ours}
         \end{subfigure}
             \caption{Harmful samples in the real world.}  
             \label{fig:real}     
 \end{figure}
\subsection{Examples of NSFW memes in Real World}
\label{app:G}
Figure \ref{fig:real} provides examples of harmful images detected on the Memedroid website. Figure \ref{fig:real_common} shows a harmful meme detected by both \system{} and the comparison methods, while Figure \ref{fig:real_ours} shows harmful memes detected only by \system{}. 
Figure \ref{fig:real_ours} includes memes with implicit sexual content (even links to pornographic content) and warning messages. These types of images were not identified by the platform or existing methods.


\end{document}